\documentclass{acmsiggraph}

\usepackage{amsmath}
\usepackage{amssymb}
\usepackage{latexsym, bm}

\newcommand{\ie}{\textit{i}.\textit{e}.}
\newcommand{\eg}{\textit{e}.\textit{g}.}

\newcommand{\Tref}[1]{Table~\ref{#1}}

\newcommand{\Fref}[1]{Figure~\ref{#1}}

\newcommand{\eref}[1]{Eq.~(\ref{#1})}
\newcommand{\fref}[1]{Fig.~\ref{#1}}

\title{Waterdrop Stereo}

\author{Shaodi You \thanks{e-mail:youshaodi@gmail.com} \\NICTA
\and Robby T. Tan \thanks{e-mail:tanrobby@gmail.com} \\ Yale-NUS College
\and Rei Kawakami \thanks{e-mail:rei@nae-lab.org} \\The University of Tokyo
\and Yasuhiro Mukaigawa \thanks{e-mail:mukaigawa@is.naist.jp} \\NAIST
\and Katsushi Ikeuchi \thanks{e-mail:katsuike@microsoft.com} \\Microsoft Research
}

\pdfauthor{Paper ID: 0526}

\TOGonlineid{45678}


\setcopyright{rightsretained}
\copyrightyear{2016}

\begin{document}

%%% This is the ``teaser'' command, which puts an figure, centered, below 
%%% the title and author information, and above the body of the content.

 \teaser{
\includegraphics[width=\linewidth]{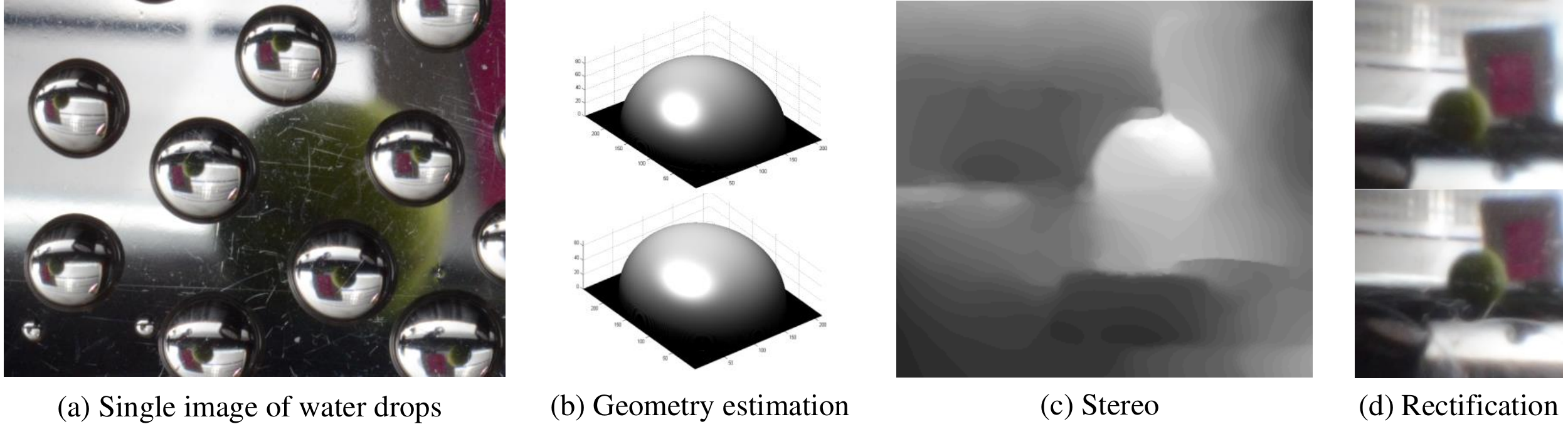}
\caption{The pipeline of the proposed method. From left to right: Input image of water drops. Estimated 3D water drop shape. Estimated depth. Geometric Rectification.}
\label{Fig:Pipeline}
 }

\maketitle

\begin{abstract}

This paper introduces depth estimation from water drops. The key idea is that a single water drop adhered to window glass is totally transparent and convex, and thus optically acts like a fisheye lens. If we have more than one water drop in a single image, then through each of them we can see the environment with different view points, similar to  stereo. To realize this idea, we need to rectify every water drop imagery to make radially distorted planar surfaces look flat. For this rectification, we consider two physical properties of water drops: (1)  A static water drop has constant volume, and its geometric convex shape is determined by the balance between the tension force and gravity. This implies that the 3D geometric shape can be obtained by minimizing  the overall potential energy, which is the sum of the tension energy and the gravitational potential energy. (2) The imagery inside a water-drop is determined by the water-drop 3D shape and total reflection at the boundary. This total reflection generates a dark band commonly observed in any adherent water drops. Hence, once the 3D shape of water drops are recovered, we can rectify the water drop images through backward raytracing. Subsequently, we can compute depth using stereo. In addition to depth estimation, we can also apply image refocusing. Experiments on real images and a quantitative evaluation show the effectiveness of our proposed method. To our best knowledge, never before have adherent water drops been used to estimate depth.

\end{abstract}


%\ccsdesc[500]{Computing methodologies~Image manipulation}
%\ccsdesc[300]{Computing methodologies~Computational photography}


% The next three commands are required, and insert the user-generated keywords, 
% The CCS concepts list, and the rights management text.
% Please make sure there is a blank line between each of these three commands.

\keywordlist

\conceptlist

\printcopyright

\begin{figure*}[tb]
	\includegraphics[width=\linewidth]{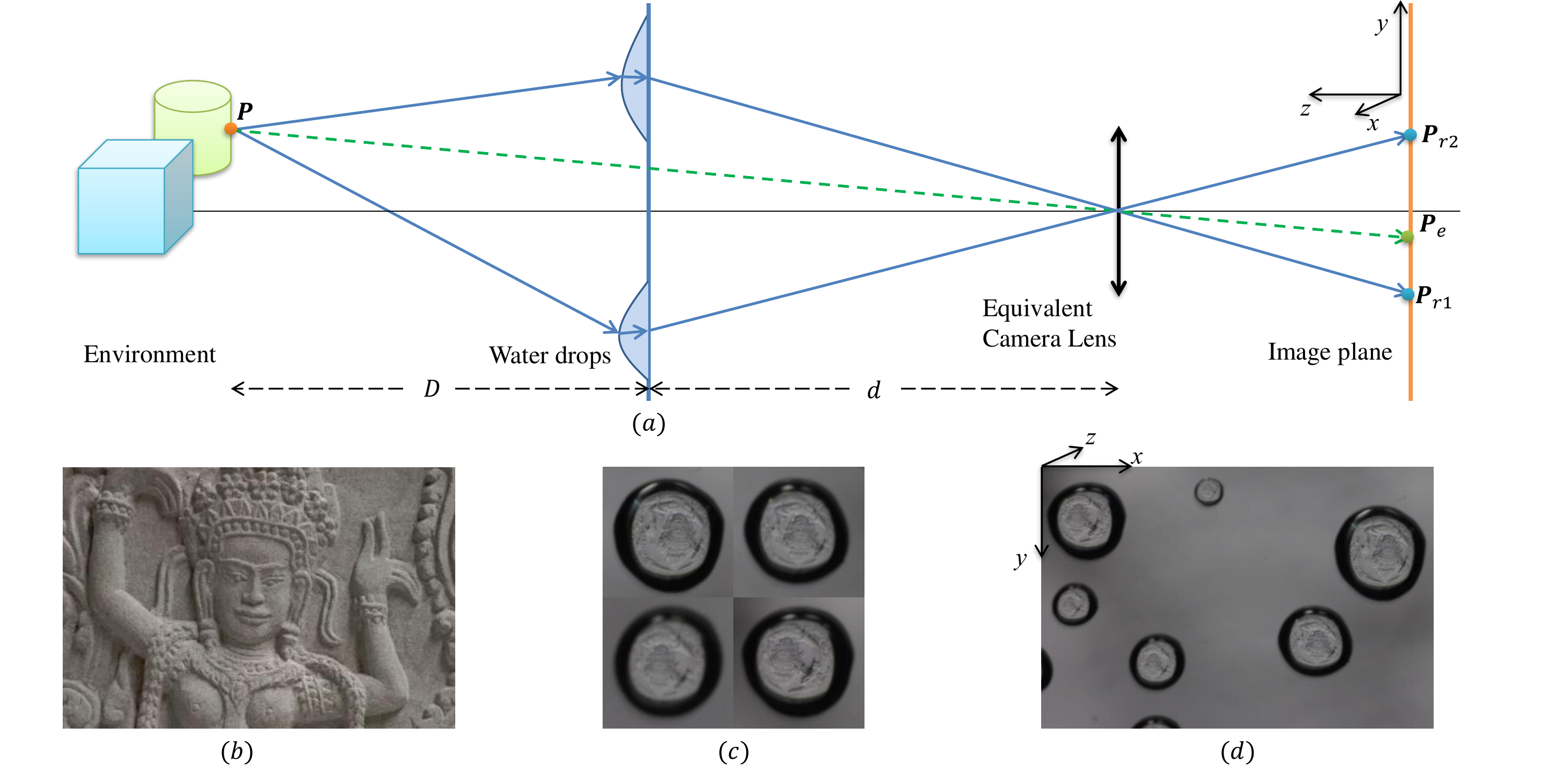}
	\caption{Image formation.} (a) The light path model assuming a pinhole camera. (b) Appearance of the environment when the camera focuses on the environment. (c) Appearance of water drops. (d) Image obtained by the camera.
	\label{Fig:PinHole}
\end{figure*}

\section{Introduction}
Depth from real images is crucial information for many applications in computer graphics. A numerous methods   have attempted to extract depth with various cues ( e.g., shading \cite{Ikeuchi81}, multi-views \cite{Hartley04}, defocus \cite{Favaro05}). In contrast to all existing methods, in this paper, we explore a new possibility of using water drops adhered to window glass or a lens to estimate depth.

Water drops adhered to glass are  transparent and convex, and thus each of them acts like a fisheye lens.
As shown in \fref{Fig:Pipeline}.a, water drops' locations are normally scattered in various regions in an image. If we zoom in, each of the water drops displays the same environment from its own unique point of view. Due to the proximity to each other, some have similar visual content, but some can be relatively different, particularly when the water drops are apart in the image.
Therefore, if we can rectify each of the water drops, we will have a set of images of the environment from relatively different perspectives, opening up the possibility of extracting the depth from the water drops, which is the goal of this paper.

To be able to achieve the goal, we need to rectify each water drop, so that planar surfaces look flat.  Rectifying water drops, however, is problematic. In contrast to  existing work in catadioptic imaging, which assumes the geometry of the sphere is known a priori, water drops shapes can vary in a considerable range and highly non-axial. To resolve this problem, we need to examine two physical properties of water drops. First, a static water drop has constant volume, and its geometric 3D shape is determined by the balance between the tension force and gravity.  Because the water drop is in balance, it minimizes the overall potential energy, which is the sum of the tension energy and the gravitational potential energy. Based on this property, we introduce an iterative method to form the water-drop geometric shape.
However, from a single 2D image, the volume cannot be directly obtained, since we do not know the thickness of the water drop. To solve this, we use the second physical property, i.e., water-drop appearance depends on their geometric shape and also the total reflection. The total reflection occurs near the water drop boundaries and triggers a dark band. We found that a water drop with a greater volume will have a wider dark band. Thus, we introduce  a volume-varying-iteration framework that estimates the volume that best fit to the appearance. Having known the complete 3D shape of water drops, we perform the multiple view stereo through backward raytracing and triangulation. Finally, we rectify the warped images. Having obtained the depth and rectified images, one of our applications is image refocusing. \Fref{Fig:Pipeline} shows the general pipeline of our proposed method. 

\paragraph{Contributions} In this paper, we introduce a new way to recover depth using water drops from a single image. We also propose a novel method to reconstruct the 3D geometry of water drops  by utilizing the minimum surface energy and  total reflection. Aside from estimating depth, we also apply image refocusing through the information provided by water drops. Furthermore, the proposed non-parametric, non-axial algorithm can be generally applied to catadioptic imaging system.

The rest of the paper is organized as follows. Section 2  discusses related work  in depth estimation, water modeling and shape from transparent objects. Section 3 explains the theory behind the water-drop physical properties. Section 4 introduces the  methodology of the 3D shape estimation, stereo, as well as water-drop image rectification. Section 5 shows the three applications on stereo, image refocusing and image stitching. Section 6 shows the experimental results and evaluation. Section 7 concludes this paper.

%%%%%%%%%%%%%%%%%%%%%%%%%%%%%%%%%%%%%%%
%Section
%%%%%%%%%%%%%%%%%%%%%%%%%%%%%%%%%%%%%%%
\section{Related Work}

Three dimensional reconstruction of opaque objects from a single image have been explored for decades:  Shape from shading \cite{Ikeuchi81}, shape from texture \cite{Malik97}, shape from defocus \cite{Favaro05} and piece-wise planarity \cite{Horry97}. A few approaches using silhouettes have been proposed  to reconstruct  a bounded smooth surface \cite{Terzopoulos88,Hassner06,Prasad06,Joshi08,Oswald12,Vicente13}. The method of \cite{Prasad06} reconstructs a surface with minimum area, and \cite{Oswald12} proposes its speed-up version. However, none of these methods directly aim to model water or other transparent liquid from a single image.

Methods of \cite{Garg07,Roser09,You13,You15} introduce airborne and adherent raindrop modeling. Their goal is to detect and remove raindrops, and not to reconstruct 3D structure from raindrops. \cite{Roser10} exploits  water drop surface fitting using B-splines and silhouettes using 1D splines.
\cite{Morris04,Tian09,Oreifej11,Kanaev12} exploit underwater imaging. They assume  water surfaces are dynamic and  dominated by transitions of waves, which do not suit to our specific problem.

Stereo and light field using perspective cameras with extra mirrors and lenses have also been explored. \cite{Baker99,Taguchi10} propose algorithms using sphere mirrors. \cite{Levoy04} introduces  arrays of planar mirrors. \cite{Swaminathan01,Ramalingam06} address the use of axial cameras, and \cite{Taguchi10} extends the work to axial-camera arrays. Later, \cite{Agrawal13} proposes methods to automatically calibrate the system.
All of these methods, however, assume  radial or planar symmetry of the media (mirror/lens), which are not satisfied in the case of water drops, since water drops are highly non-axial.

%%%%%%%%%%%%%%%%%%%%%%%%%%%%%%%%%%%%%%%
%Section
%%%%%%%%%%%%%%%%%%%%%%%%%%%%%%%%%%%%%%%
\section{Modeling}

Theoretical background and modeling of water drops are discussed in this section. We first explain briefly the image formation, showing the correlations between the environment, water-drops and the camera. Subsequently, we model the raindrop 3D geometry, particularly the concept of minimum energy surface. Based on the image formation and the raindrop geometry, we study the total reflection inside water drops, which is necessary to determine the water-drop's volume.  All these aim at water-drop image rectification.

\subsection{Image Formation}

\fref{Fig:PinHole} illustrates our image formation. Rays reflected from the environment pass through two water drops before hitting the image plane. Unlike in the conventional image formation, the passing rays are refracted by water drops, where each water drop acts like a fisheye lens that warps the images. Assuming we have a few water drops that are apart to each other,  the imageries of the water drops will be slightly different to each other, even though the environment is identical, as shown in \fref{Fig:PinHole}.

From the illustration in \fref{Fig:PinHole}, we can conclude that the image captured by the camera through water drops is determined by three interrelated factors: (1) the depth of the environment, (2) the 3D shape of water drops, which determine how light rays emitted from the environment are refracted and, (3) camera intrinsic parameters, which are assumed to be known.
Therefore, to be able to recover the depth of the environment, we need to obtain the 3D shape of water drops.

\subsection{Minimum Energy Surface}

\begin{figure}[tb]
	\includegraphics[width=\linewidth]{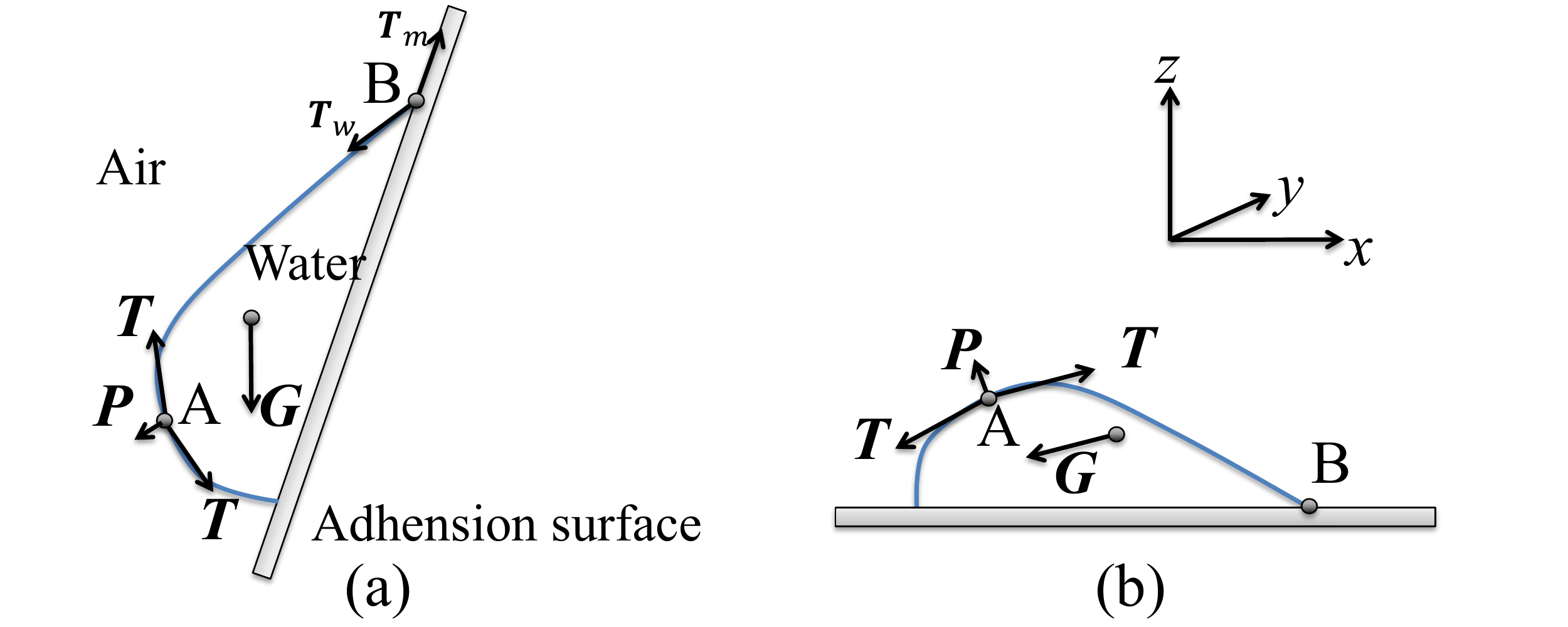}
	\caption{Parameters of a water drop.} (a) In the global coordinates, the geomertic shape is determined by the gravity. (b) The parameters in the camera coordinates. Point A is a two phase point (water-air), where the tensor force balances the pressure. Point B is a three phase point (water-air-material).
	\label{Fig:TensorEnergy}
\end{figure}

%Unfortunately, the tension of a three phase point is highly %relying on the material of the adhesion surface, which in our case is the glass.
%Alternatively, we look into the overall potential energy of the system. We utilize %the property of minimum energy surface. Based on it, we can solve the raindrop %geometry by only inferring the adhesion area ($\Omega_R$) and the volume ($V$) %from the image.

%\textbf{Minimum energy surface.}

For the purpose of exploring the minimum energy surface to estimate the 3D shape of a water drop, we  introduce a local coordinate system of the camera, which is illustrated in \fref{Fig:PinHole}.a~and~d. In the coordinates, water drop 3D shape can be parameterized as:
\begin{equation}
\mathcal{S} = \{ z(x, y), (x, y)\in\Omega_R\},
\label{Eq:Surface}
\end{equation}
where $\Omega_R$ indicates the raindrop area attached to glass. $(x, y)$ is any point in the raindrop area and $z$ is the height.

A static water drop has a constant volume, and its 3D shape $\tilde{\mathcal{S}}$ minimizes the overall potential energy $E$, which can be written as:
\begin{equation}
\begin{split}
E({\tilde{\mathcal{S}}}) &= \min{E}(\mathcal{S}) = \min({E}_T(\mathcal{S}) + E_G(\mathcal{S}) ),
\\
V(\mathcal{S}) &= constant,
\end{split}
\label{Eq:Engery}
\end{equation}
where ${E}_T$ is the tension energy, and ${E}_G$ is the gravitational potential energy, $V$ is the volume. Therefore, to solve the geometry of a raindrop, we need to find the surface $\tilde{\mathcal{S}}$.

\Fref{Fig:TensorEnergy} illustrates the 3D shape of a water drop.
Point $A$ is a two-phase (water-air) balanced point,  where  surface tension $\bm{T}$ balances  pressure $\bm{P}$. Point B is a three phase point (water-air-material), where the tension is from both water $\bm{T}_w$ and adhesion surface $\bm{T}_m$. These two types of tension balance the gravity, $\bm{G}$.

With the parameterized surface, we can write the surface tension energy as:
\begin{equation}
E_T(\mathcal{S}) = \int_{\Omega_R}{\sigma\mathrm{d}A} = \int_{\Omega_R}{\sigma\sqrt{1 + |\nabla{z}|^2 }\mathrm{d}x\mathrm{d}y},
\label{Eq:Tension}
\end{equation}
where $\sigma$ is the surface tension index for water, $\mathrm{d}A$ denotes a unit surface area and $\nabla$ is the gradient \cite{Feynman13}. As we can see, the tension energy is proportional to the area of the surface.

\begin{figure}[tb]
	\includegraphics[width=\linewidth]{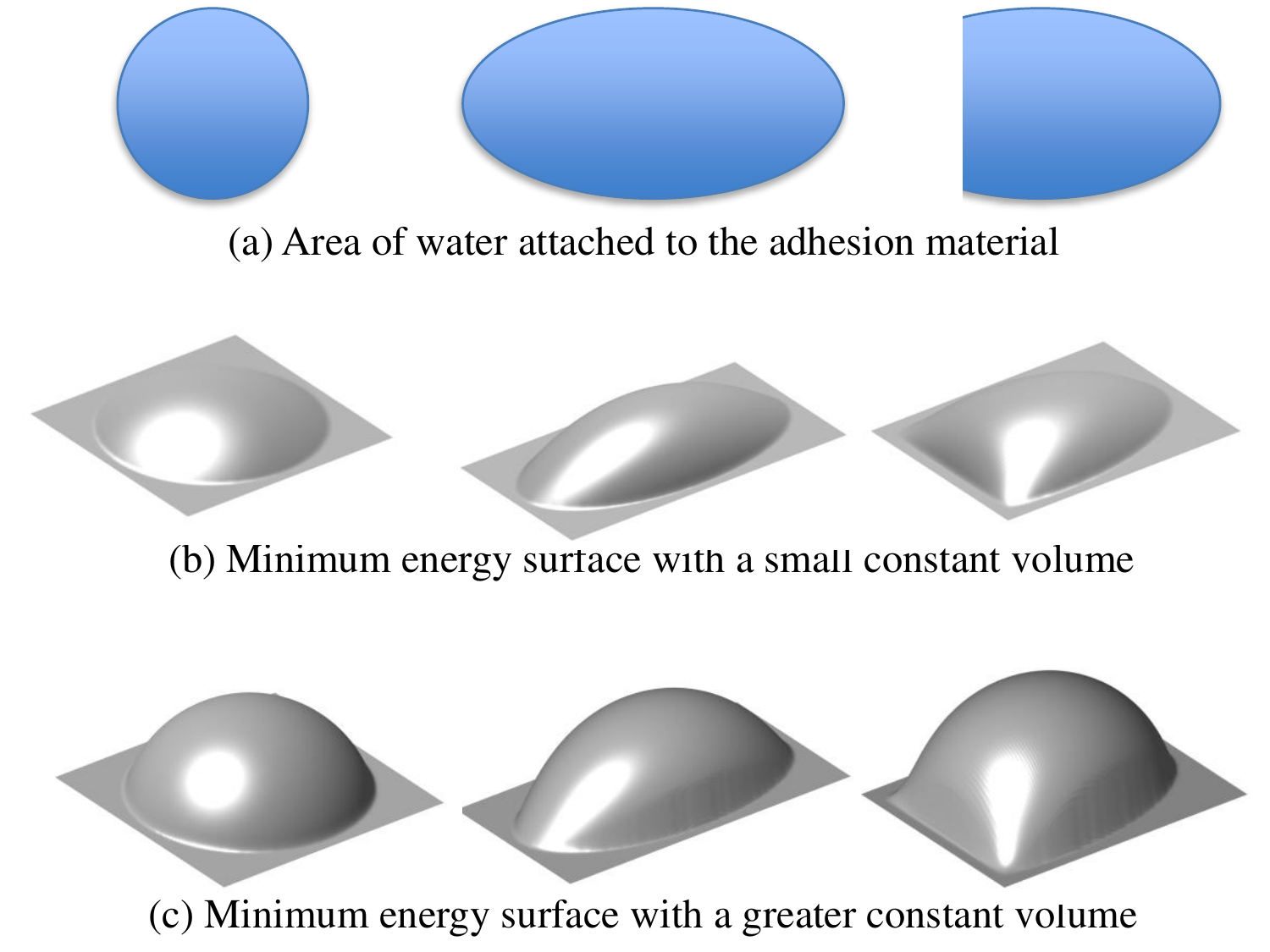}
	\caption{Minimum energy surfaces given the area and volume.} (b) The minimum energy surface when the volume coefficient $\alpha = 0.10$, defined in \eref{Eq:Volume2}. (c) $\alpha = 0.35$. Assuming the gravity is along $z$ axis.
	\label{Fig:Volume}
\end{figure}

The gravitational potential energy can be expressed as:
\begin{equation}
E_G(\mathcal{S}) = \int_{\Omega_R}{\mathrm{d}x\mathrm{d}y}\int_{0}^{z}{(x\cos{\theta_x}+y\cos{\theta_y}+z\cos{\theta_z})g\rho\mathrm{d}w},
\label{Eq:Gravity}
\end{equation}
where $\theta_x$, $\theta_y$ and $\theta_z$ denote the angles between the $x, y, z$ coordinates and the gravity correspondingly. $g$ is the gravity and $\rho$ is the density of water, which are generally known.
Moreover, we can add a constraint that:
\begin{equation}
\int_{\Omega_R}{z\mathrm{d}x\mathrm{d}y}\equiv{V}.
\label{Eq:Volume}
\end{equation}
Therefore, the parameterized surface $\mathcal{S}$ is estimated by minimizing the overall potential energy determined by \eref{Eq:Engery}, (\ref{Eq:Tension}) and (\ref{Eq:Gravity}) with the constraints of constant volume in \eref{Eq:Volume}.
\Fref{Fig:Volume} shows some examples of the surface estimated by using the technique.
We will discuss the algorithm in detail in Section 4. 

Note that, to uniquely determine the geometry of a water drop, we need to know both the 2D area where the water drop attached to glass, $\Omega_R$, and the volume $V$. While the former can be directly inferred from the image, the latter is not straightforward to obtain. The subsequent section will discuss how we can possibly determine the volume.

\subsection{Water-Drop Volume from Dark Band}
As we can see in \fref{Fig:PinHole}.c, the basic idea of our volume estimation is based on the dark band at the boundary of a water drop. We found that the wider the dark band the larger the volume of the water. This section discusses this idea further.

\begin{figure*}[tb]
	\includegraphics[width=\linewidth]{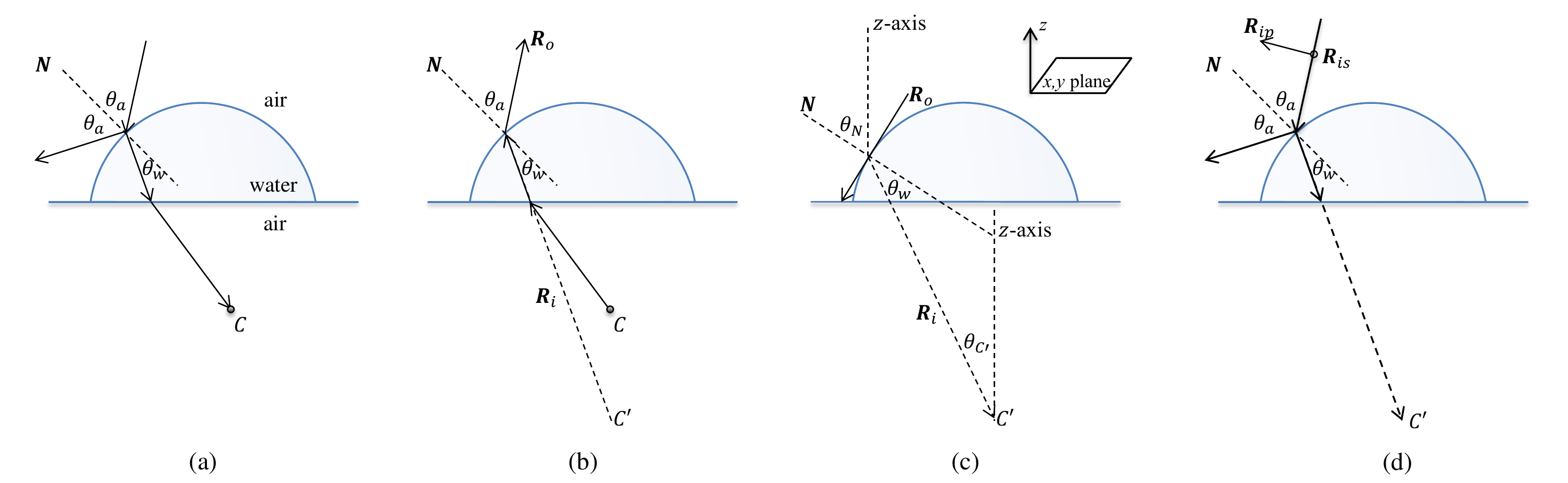}
	\caption{Refraction by a water drop.} (a) A ray coming from the environment is refracted twice before reaching the camera $C$. (b) Backward ray tracing, where a virtual ray emitted from the camera passes through the same path as in (a). For simplification, we remove the refraction between flat surfaces by moving the camera position to $C'$. (c) When $\theta_w$ is greater than the critical angle, light will not be transmitted but reflected inside. (d) Two polarized components, $R_s$ and $R_p$, of the incidence ray.
	\label{Fig:Brewster}
\end{figure*}

\paragraph{Refraction model}
\fref{Fig:Brewster}.a illustrates a ray coming from the environment is refracted twice before reaching the camera. Since, we are only interested in the rays that can reach the camera, we can use backward raytracing to know the paths of the rays. Moreover, we assume that the glass is so thin that we can ignore the refraction due to the glass.
Because we are mostly interested in the refraction on the curved surface, to further simplify the model, we remove the refraction between the flat surface by moving the camera from position $C$ to $C'$, as shown in \fref{Fig:Brewster}.b. 
For approximation, when the incident angle is small, we can consider that the perpendicular distance from the camera to the refraction plane, denoted as $C_z$, is changed to $\hbox{$C'_z = \frac{n_w}{n_a}C_z$}$, where $n_w$ and $n_a$ are the refractive indices of water and air, respectively. 
Detailed derivation of the position of $C'$ is discussed in Appendix A.

\paragraph{Dark Band and Total Reflection}
The dark band at the boundary of a water drop
is caused by light coming from the environment  reflected back inside the water, instead of being transmitted to the camera. This phenomenon is known as the total reflection, and applies to all light rays whose relative angles  to the water's surface normal are larger than the critical angle, denoted as $\theta_W$.

To analyze the correlation between the critical angle with the water-drop 3D shape \hbox{$S : z(x, y)$}, we refer to  Snell's law, which indicates the critical angle:
\begin{equation}
\tilde{\theta}_{w} = \sin^{-1}\frac{n_a}{n_w}.
\label{Eq:Snell2}
\end{equation}
As indicated in \fref{Fig:Brewster}.c, we denote the surface normal as $\mathbf{N}$, which can be derived from $z$ as:
%\begin{equation}
$
%\begin{split}
\mathbf{N} = (N_x, N_y, N_z)^\top
= \frac{\mathbf{N'}}{\parallel\mathbf{N'}\parallel}
$
where,
$
\mathbf{N}' = (\frac{\partial{z}}{\partial{x}}, \frac{\partial{z}}{\partial{y}}, 1)^\top,
%\end{split}
$
%\label{Eq:Normal}
%\end{equation}
and $\parallel~~\parallel$ denotes the $\ell_2$ norm.

\begin{figure}[tb]
	\includegraphics[width=\linewidth]{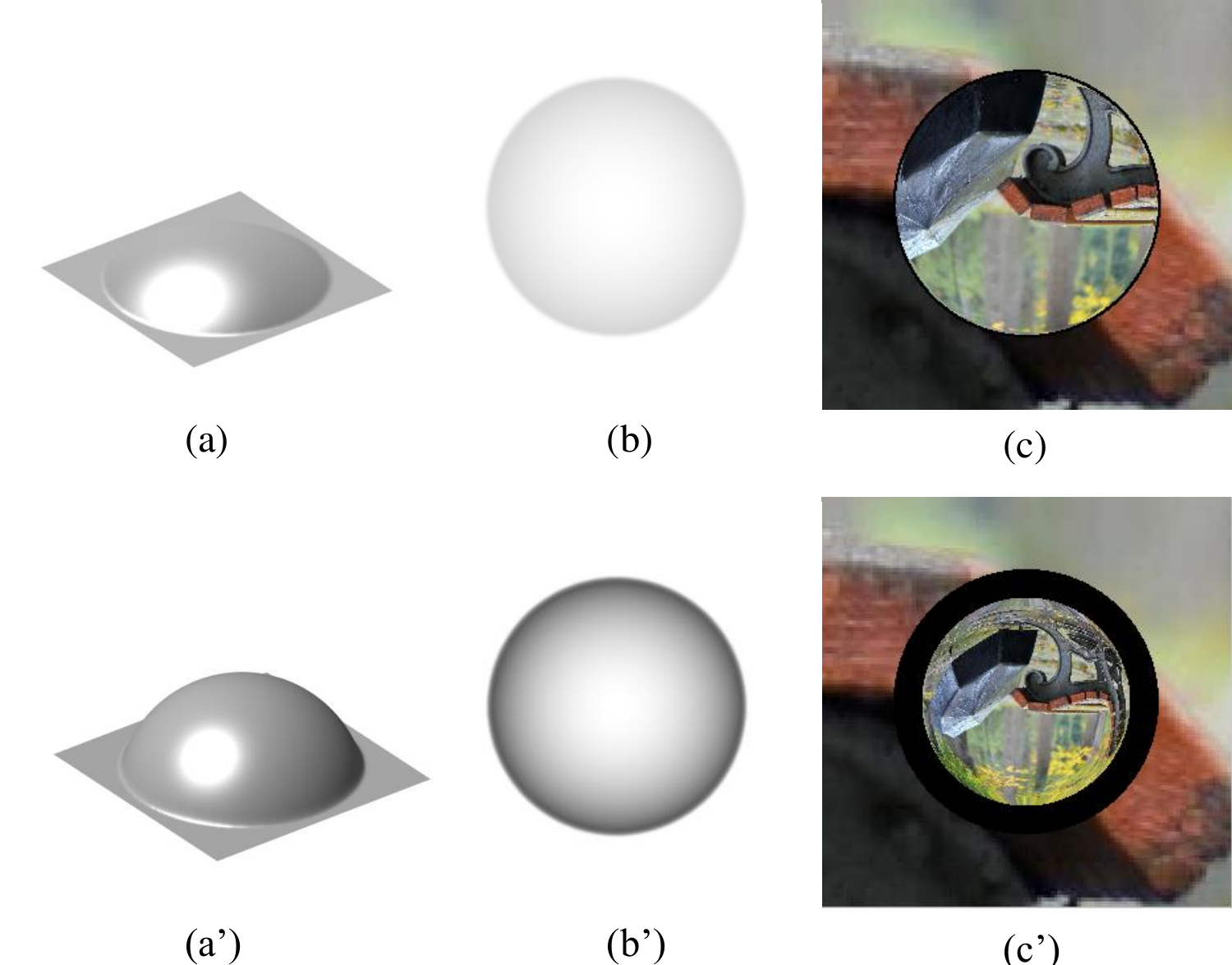}
	\caption{Dark bands.} (a) Water drop 3D geometry. (b) The $z$ component of surface normal. (c) The  dark bands. Second row: A greater water drop volume, wider dark band.
	\label{Fig:DarkBand}
\end{figure}

The angle between the surface normal and the $z$-axis  denoted as $\theta_N$ is the sum of the incidence angle of water, $\theta_w$, and the angle between the incidence ray and $z$-axis $\theta_{C'}$:
\begin{equation}
\theta_N = \theta_w + \theta_{C'}.
\label{Eq:Normal2}
\end{equation}
where $\theta_{C'}$ is determined by the position of the camera and the position of the refraction. Considering the $z$ component of the normal $N_z$, also defined as: $N_z = \cos\theta_N$, we know that when \hbox{$N_z \leq \cos( \tilde{\theta}_w + \theta_{C'})$}, the corresponding water drop area is totally dark.
For instance, when $\theta_{C'}$ is 0, and $\frac{n_a}{n_w}$ is approximately $\frac{3}{4}$, we have 
\begin{equation}
N_z \leq \tilde{N}_z \approx 0.661.
\label{Eq:Normal4}
\end{equation}
Where $\tilde{N}_z$ is the denotation for the critical value.
\Fref{Fig:DarkBand} shows some examples of synthetically generated dark bands.
As we can observe, a greater volume of the water drop indicates a wider dark band. Therefore,  to infer the water drop volume from the dark band is possible.

\begin{figure*}[tb]
	\includegraphics[width=\linewidth]{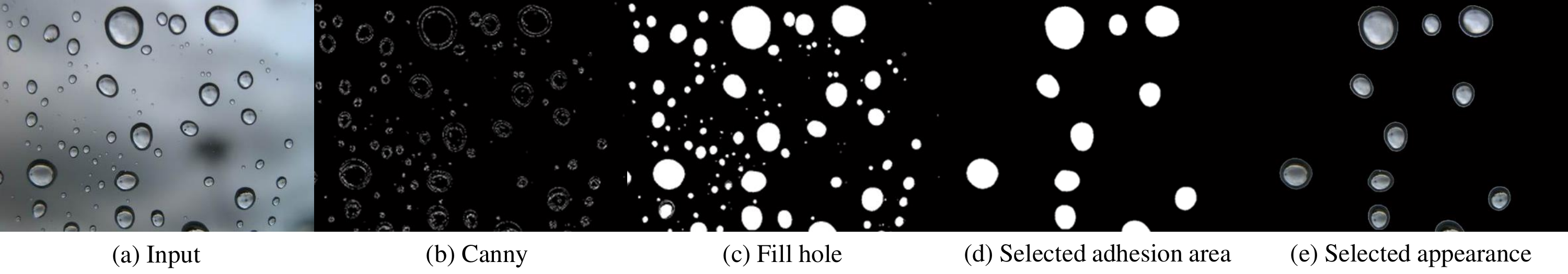}
	\caption{Selecting water drops from a single image.}
	\label{Fig:Canny}
\end{figure*}

\begin{figure*}[tb]
	\includegraphics[width=\linewidth]{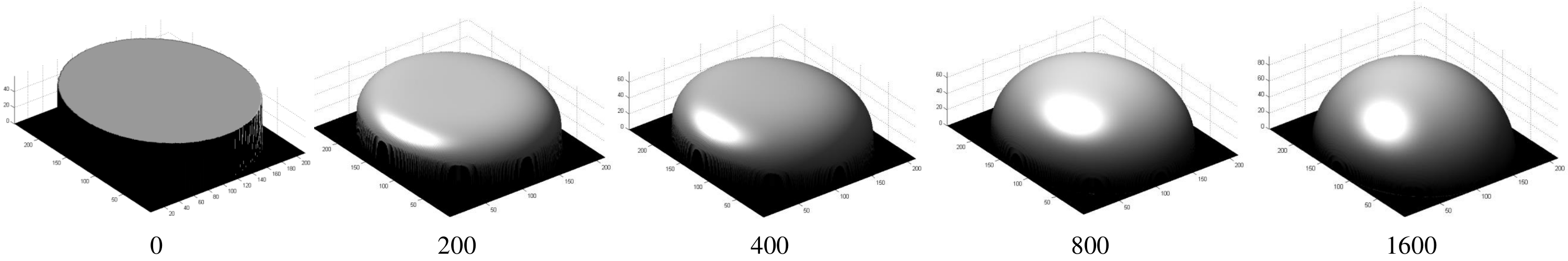}
	\caption{Iteration of water drop 3D shape with a fixed volume.}
	\label{Fig:Iteration}
\end{figure*}

%%%%%%%%%%%%%%%%%%%%%%%%%%%%%%%%%%%%%%%
\paragraph{Dark Band and Fresnel Equation }
While the dark band can be theoretically inferred from the water-drop geometry, detecting them from an image is nontrivial. Due to the sensor noise and the leak of light\footnote{Back light from camera side which is reflected by the glass plate and goes into the camera; and the interreflection inside a water drop.}, dark bands are not totally dark. Moreover, there are textures in the environment that can be darker than dark bands. To resolve the problem, we employ the Fresnel equation and formulate the brightness values near the critical angle.

The refraction coefficients, denoted as $\mathcal{T}_s$ and $\mathcal{T}_p$, for two orthogonal polarized components for the light rays traveling from air to water are written as:
\begin{equation}
\mathcal{T}_s = 1 - \left(\frac{\sin(\theta_w-\theta_a)}{\sin(\theta_w+\theta_a)}\right)^2
\label{Eq:Ts}
\end{equation}
\begin{equation}
\mathcal{T}_p = 1 - \left(\frac{\tan(\theta_w-\theta_a)}{\tan(\theta_w+\theta_a)}\right)^2
\label{Eq:Tp}
\end{equation}
where $\theta_w$ and $\theta_a$ are depicted in Fig.~\ref{Fig:Brewster}.d.
In our case, we assume the light from the environment is not polarized, and thus the overall refraction coefficient is
\begin{equation}
\mathcal{T} = \frac{1}{2} (\mathcal{T}_s + \mathcal{T}_p).
\label{Eq:T}
\end{equation}

Concerning the dark bands, we are interested in two critical conditions. First, when the incidence angle $\theta_a$ is close to 0. In such a condition, $\sin\theta_a\approx\theta_a$, \hbox{$\cos\theta_a\approx1$}, and consequently:
\begin{equation}
\mathcal{T} = \frac{4n_an_w}{(n_w + n_a)^2}.
\label{Eq:T2}
\end{equation}
Substituting the value for water gives us $\frac{n_a}{n_w} = \frac{3}{4}$, and thus we have $\mathcal{T}  \approx 0.980$.

Second, when incidence angle $\theta_a$  is close to $\frac{\pi}{2}$, (the locations near the dark band). Hence, \hbox{$\sin\theta_a\approx1$}, \hbox{$\cos\theta_a\approx \frac{\pi}{2} - \theta_a$}, as a result:
\begin{equation}
\mathcal{T} = 2 \sqrt{ 1 - \left(\frac{n_a}{n_w}\right)^2}\left(\frac{n_a}{n_w} + \frac{n_w}{n_a}\right)\left( \frac{\pi}{2} - \theta_a\right)
\label{Eq:T3}
\end{equation}
Similar to the first condition, substituting the value $\frac{n_a}{n_w} = \frac{3}{4}$, we obtain:
\begin{equation}
\mathcal{T}  \approx 2.76(\frac{\pi}{2} - \theta_a).
\label{Eq:T4}
\end{equation}
Considering $\theta_a$ is connected with $\theta_w$ by Snell's law, and the connection between $\theta_N$ and $\theta_w$ (\eref{Eq:Normal2}), we establish the connection between the $z$ component of surface normal $N_z = \cos\theta_N$ and the refraction coefficient:
\begin{equation}
\mathcal{T}  \approx 7.68(N_z - \tilde{N}_z).
\label{Eq:T5}
\end{equation}

This relation between the image brightness and surface normals give us a constraint of a contrast between the dark band and other parts inside the waterdrop image. Using this, the water-drop volume can be inferred using the brightness close to the dark band region. Details are provided in Sec. 4.1

%%%%%%%%%%%%%%%%%%%%%%%%%%%%%%%%%%%%%%%
%
%	\paragraph{Scope of water drops}
%	Water drops allows us to convert a common camera to 
%	a catadioptric camera. Referring to 
%	\fref{Fig:Brewster}.c, according to 
%	\eref{Eq:Normal2}, each of the water drop allows to 
%	scope about $ \pi - 2\theta_w \approx 83^o$. 
%	Considering the water drops are distributed in all %
%	the image, the scope of all the raindrops is 
%	extended to: $\theta_s + 83^o$, where $\theta_s$ is 
%	the original scrop of the image. This mean water 
%	drops can always extend the scrop of the image by 
%	$83^o$.

%%%%%%%%%%%%%%%%%%%%%%%%%%%%%%%%%%%%%%%
%Section
%%%%%%%%%%%%%%%%%%%%%%%%%%%%%%%%%%%%%%%
\section{Methodology}

In this section, the detailed algorithm for rectifying images of water drops and estimating depth is introduced. As illustrated in \fref{Fig:Pipeline}, it has three main steps: (1) water drop 3D shape reconstruction by minimizing energy surface, (2) Multi-view stereo and (3) Image rectification.

\paragraph{Water Drop Detection}

Water-drops appearance is highly dependent on the environment, and thus detecting them is not trivial. Fortunately, in our case, we can assume water drops are in focus, and thus the environment image is rather blurred. Hence, we can utilize edge detection to locate water drops, as illustrated in \fref{Fig:Canny}. Having located water drops, we select those that are sufficiently large (\eg, the diameter is greater than 300 pixels). This is to ensure that rectified images are not too small.

\subsection{Water Drop 3D Shape Reconstruction}

\paragraph{Mesh representation and Initialization}
To reconstruct the 3D shape of water drops, we first represent the water surface using a parameterized mesh.
Referring to \eref{Eq:Surface}, we can describe a surface as:
$
\mathcal{S} = \{ z(i, j), (i, j)\in\Omega_R\},
$
where $(i, j)$ are the location of a pixel in the water drop area.
Accordingly, the area of $\Omega_R$ is defined as:
$
B = \sum_{(i, j)\in\Omega_R}1,
$ where 1 is the unit for a pixel's area.

At this initialization, we do not know the volume of the water drop, and make an initial guess based on:
\begin{equation}
	V = \alpha B^\frac{3}{2},
	\label{Eq:Volume2}
\end{equation}
where $\alpha$ is  the volume coefficient and  set to 0.30 as default. Based on the equation, with $\alpha$ fixed, when the area $B$ increases in square rate, the volume will increase in cubic rate. This means when performing scale change for the water drop surface, $\alpha$ remains the same value. \Fref{Fig:Volume} gives some examples how $\alpha$ is related to the reconstructed surface.

We initialize the mesh as a cylinder by defining:
\begin{equation}
	z(i, j) = \alpha B^\frac{1}{2}, (i, j)\in\Omega_R.
	\label{Eq:Mesh}
\end{equation}
\Fref{Fig:Iteration} shows an example of the initial surface.

\paragraph{Iteration with fixed volume}
We solve the constrained minimum energy surface using the iterative gradient descent. For iteration $t$ we update the mesh in three steps: tensor energy update, gravity update, and volume update. This strategy is an extension of the smooth surface reconstruction proposed by \cite{Oswald12}.

\begin{itemize}

\item 
Step 1: Tension energy update. It attempts to construct the surface as smooth as possible:
\begin{equation}
	z_{t+1} = z_{t} - \tau \sigma \cdot \frac{{\mathrm{d}} E(\mathcal{S})}{{\mathrm{d}}z_t},
	\label{Eq:TUpdate}
\end{equation}
where $\tau$ controls the update speed with $\tau = 0.5$ as default, $\sigma$ is the tension coefficient in physics. We define:
\begin{equation}
	\frac{{\mathrm{d}} E(\mathcal{S})}{{\mathrm{d}}z_t} = - {\mathrm{div}} \left( \frac{1}{\sqrt{1 + |\nabla z_t |^2}} \nabla z_t \right),
	\label{Eq:TUpdate2}
\end{equation}
where ${\mathrm{div}}$ is the divergence.

In our settings, a water drop has size around $3mm$ or approximately 500 in pixel, and thus the size of a unit pixel is about $6\mu m$. Tension coefficient for water in room temperature is $73,000 N/m$.

\item
Step 2: Gravity update. It intends to increase the height for the mesh points that lower the potential energy:
\begin{equation}
	z_{t+1}(i, j) = z_{t}(i, j) - \tau \rho g((y_g - i)\cos\theta_y + (x_g - j)\cos\theta_x),
	\label{Eq:GUpdate}
\end{equation}
where $(x_g, y_g)$ is the geometry centroid of the water drop and defined as:
\begin{equation}
	x_g = \frac{1}{B}\sum_{(i, j)}{z(i,j)\cdot j}, ~~ y_g = \frac{1}{B}\sum_{(i, j)}{z(i,j)\cdot i}.
	\label{Eq:GUpdate2}
\end{equation}

Substituting $\rho = 1Kg/L$ and $g = 9.8 m/s^2$, we found when the adherent surface tilt is small, the waterdrop geometry is mainly dominated by the tension energy.

\item
Step 3: Volume update. Having updated the tension and gravity in the previous two steps, this step checks the current volume and compares it with the targeted volume V, and then re-adjusts the volume by adding the same value to all the mesh points:
\begin{equation}
	z_{t+1} = z_t + \left( \frac {V - \sum_{(i, j)\in\Omega_R}{z_t(i,j)}}{B} \right)
	\label{Eq:VUpdate}
\end{equation}

\end{itemize}

After each iteration, we check the absolute change of volume: $\sum_{(i, j)}{|z_{t+1}(i, j) - z(i, j)|}$, and set the convergence threshold to $1e-6V$ as default, where $V$ is the targeted volume. We run the iterations up to 4000 times. \Fref{Fig:Iteration} shows the progress of the estimated volume.

\paragraph{Iteration with varying volume}

\begin{figure}[tb]
	\includegraphics[width=\linewidth]{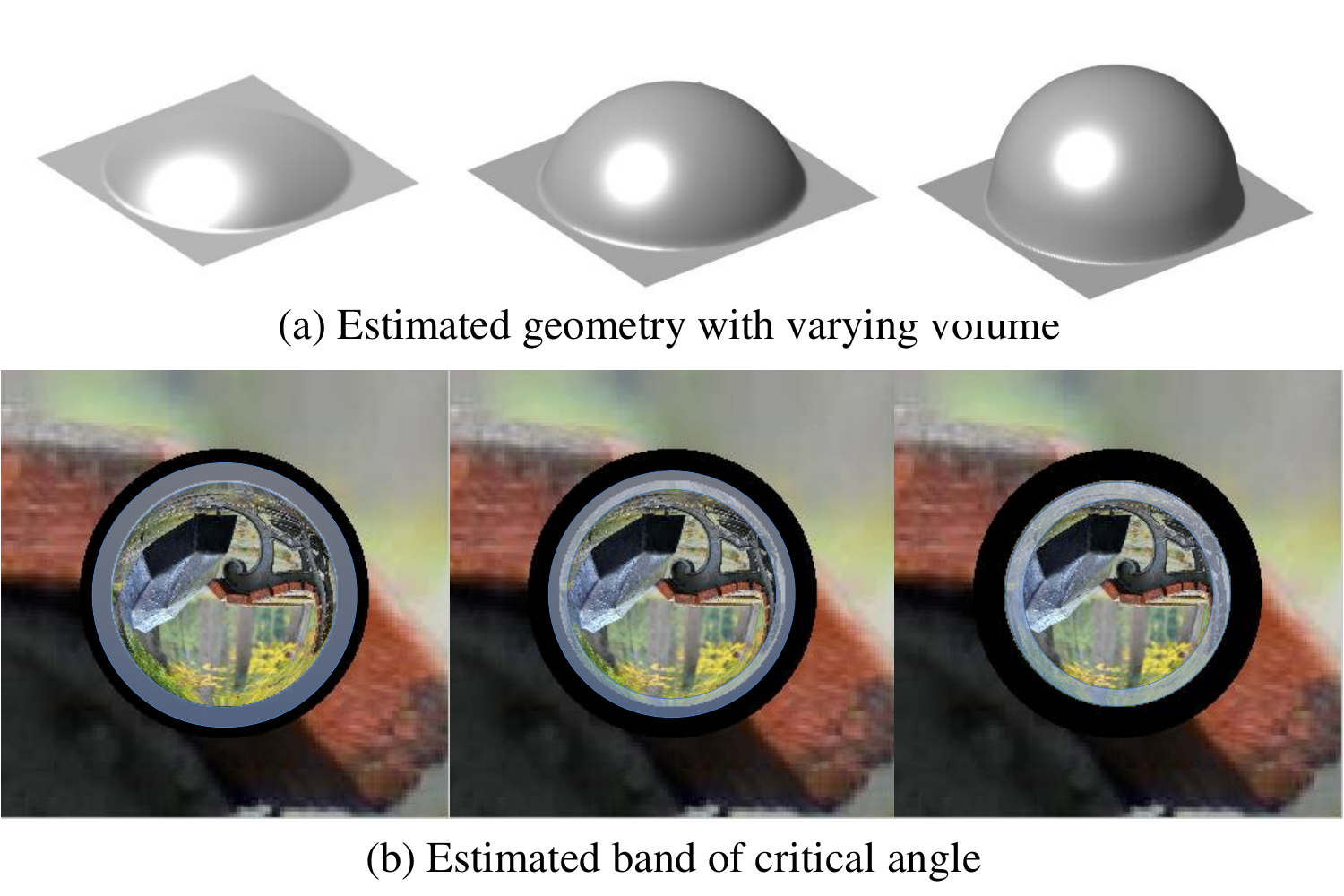}
	\caption{Estimating the width of a dark band} Left: Underestimated volume where the darkband(shown in blue) is mostly covering the dark pixels. Middle: Correctly estimated volume, where the darkband cover about the same proportion of dark and bright pixels. Right: Overestimated volume, where the darkband is mostly covering bright pixels.
	\label{Fig:Ring}
\end{figure}

Having estimated the surface with fixed volume, we can obtain the surface normals and evaluate the brightness values near the dark band and gradually adjust the volume. 

In Section 3.3, we have built the relation between the surface normal and the luminance (\eref{Eq:T5}).  According to \eref{Eq:Normal4}, when the $z$ component of surface normal is smaller than the critical value $N_Z$, the pixel brightness should be close to 0. Yet, when $z$ is slightly greater than $N_Z$, the pixel brightness should follow \eref{Eq:T5}. Thus, if we compute the average refraction coefficient, denoted as $\mathcal{T}_r$, for pixels whose normals are within the range $N_z = N_Z \pm \tilde{N}_z$, we can do local linear expansion of \eref{Eq:T4} and obtain:
\begin{equation}
T_r = 3.84 \tilde{N}_z.
\label{Eq:VUpdate1}
\end{equation}
Specifically, we set $\tilde{N}_z = 0.02\pi$, and thus $T_r = 0.241$. Consequently, the average brightness of the band, $I_r$, is:
\begin{equation}
	I_r = 0.241 I_b,
	\label{Eq:VUpdate2}
\end{equation}
where $I_b$ is the average brightness of the non water-drop areas.

In \fref{Fig:Ring}, we sample the brightness of the estimated band. As shown in \fref{Fig:Ring}.a, when the volume is underestimated, the dark band is wider than the real one, resulting less bright pixels. On the contrary, when the volume is overestimated, the dark band is narrower than the real one, resulting in brighter pixels. With the above analysis, we update the volume every 400 iteration (as default) for the fixed volume algorithm introduced previously:
\begin{equation}
	V_{t+1} = V_t + \tau_r \cdot V_{t} \cdot ( 1 -  \frac{I_t}{I_r}),
	\label{Eq:VUpdate3}
\end{equation}
where $I_t$ is the sampled brightness, $I_r$ is the targeted brightness value, and $\tau_r$ is a weighting coefficient which controls the updating speed and is set to 0.5 as default. We demonstrate the accuracy of the estimation by experiments in Sec. 5.1.

\begin{figure*}[tb]
	\includegraphics[width=\linewidth]{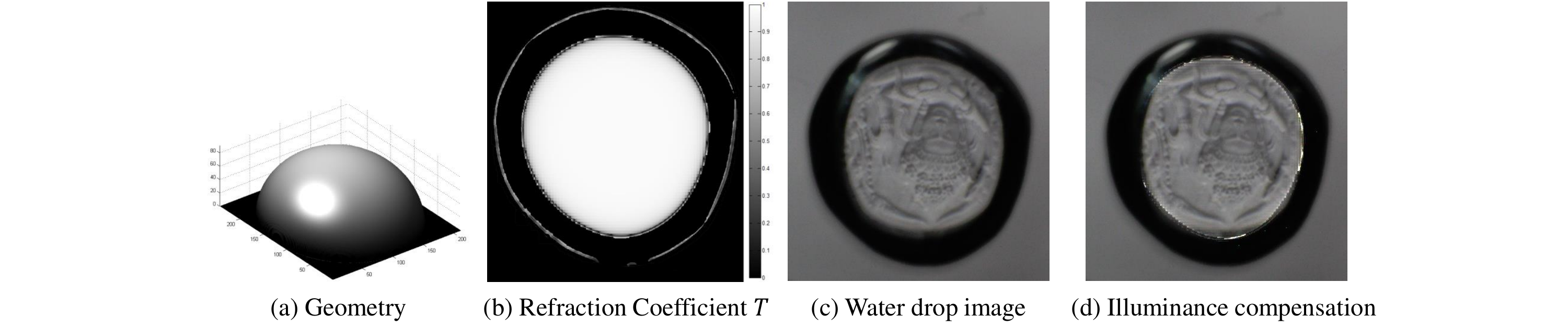}
	\caption{Illuminance compensation of water drop images.}
	\label{Fig:Illuminance}
\end{figure*}

\subsection{Waterdrop Stereo}
%\paragraph{Multi-view Stereo}
Once the geometry of each water drop is obtained, we perform multi-view stereo to estimate depth. Unlike multiple view stereo based on perspective or radial catadioptric cameras, where the projection of each camera can be modeled using a few parameters, unfortunately the water drops are non-parametric and non-axial. To overcome this problem, we propose a raytracing based triangulation. (More detail about the implementation could be found in Appendix B.)

\begin{figure}[t]
	\includegraphics[width=\linewidth]{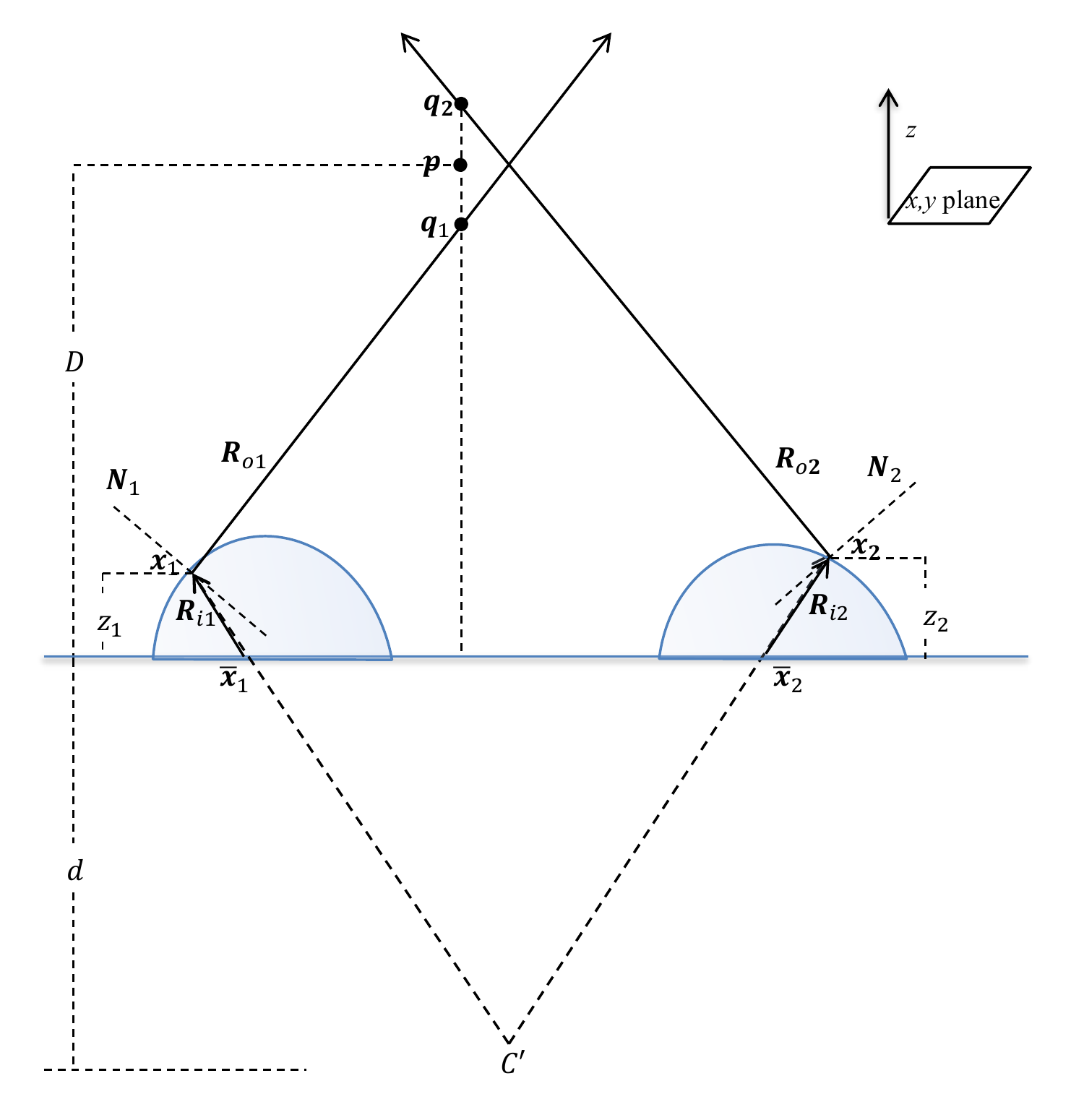}
	\caption{Ray-tracing based triangulation.}　For a set of corresponding points $\bm{x}_j$, the backward raytracing finds the set of rays $\bm{R}_{oj}$ in the space. The position $\bm{p}$ of the point in space is where the sum of Euclidean distance to all the rays is minimized.
	\label{Fig:StereoModel}
\end{figure}

As illustrated in \fref{Fig:StereoModel}, a point $\bar{\bm{x}}_1$ has its corresponding points on other water drops, denoted as $\bar{\bm{x}}_j$, where $j$ is the index of water drops. By knowing the geometry of each water drop, we can find the location where the refraction happens in each water drop, denoted as $\bm{x}_j$. 
% %
Specifically, to obtain the location of $\bm{x}$, we need: (1) The position of the camera $\bm{C}$, which is known a priori, and thus can be converted to equivalent position $\bm{C}'$, as illustrated in \fref{Fig:Brewster}.b.  (2) The water drop geometry, which is already estimated.

At each refraction location, the incident angle is obtained using \hbox{$\bm{R}_i = \frac{\bm{x} - \bm{C}'}{\parallel\bm{x} - \bm{C}'\parallel}$}. The surface normal $\bm{N}$ is known through geometry estimation. Through Snell's law, we can obtain the outbound angle $\bm{R}_o$. The outbound ray is formulated as: 
\begin{equation} 
\bm{x} + \alpha\bm{R}_o, \alpha\in\mathbb{R}.
\label{Eq:RayTracing1}
\end{equation}

\begin{figure*}[t]
	\includegraphics[width=\linewidth]{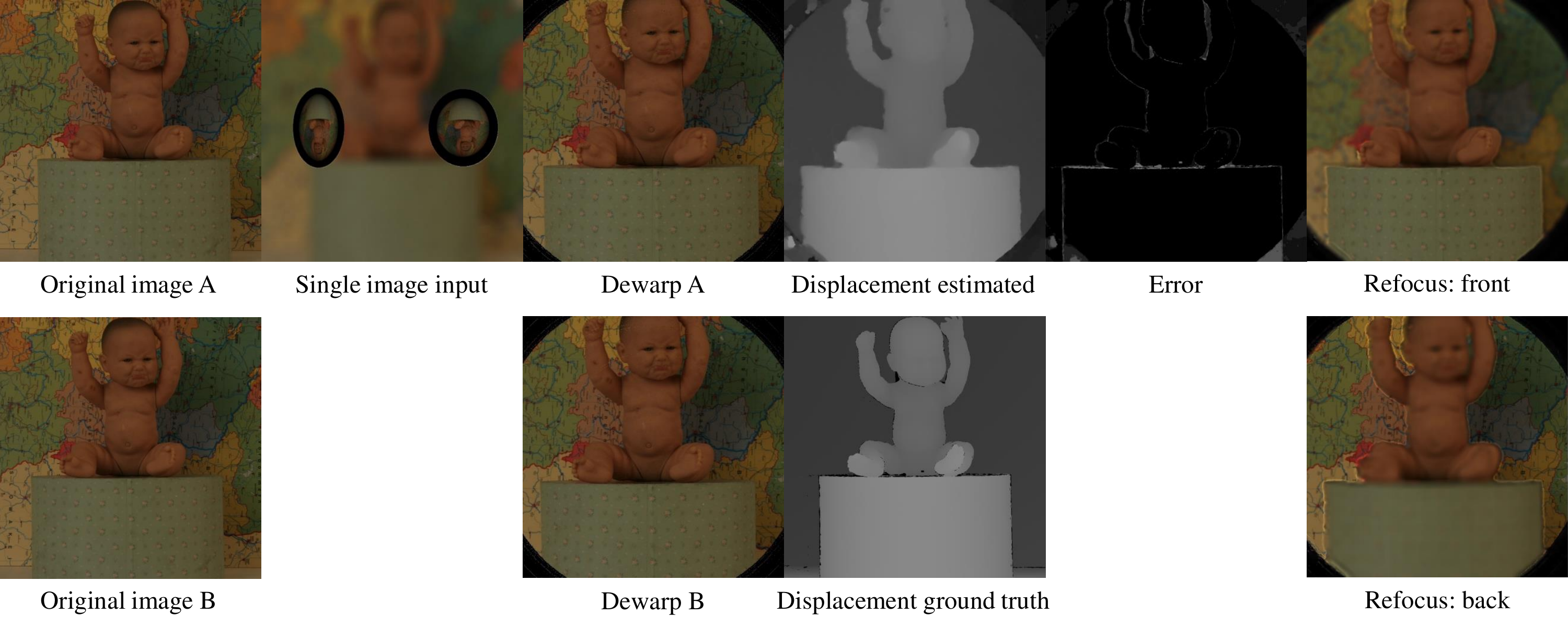}
	\caption{Stereo using two dewarped water drop images.}
	\label{Fig:2Stereo}
\end{figure*}

Hence, now we can perform the classical triangulation as illustrated in \fref{Fig:StereoModel}. Given a set of corresponding points on each waterdrop $\bm{x}_j$, we could obtain its outbound ray of fraction \hbox{$\bm{x}_j + \alpha\bm{R}_{oj}, \alpha\in\mathbb{R}, j =$} the index of water drops. The triangulation aim to find the position of point $\bm{p}$, which minimizes the Euclidean distance to all the rays:
\begin{equation} 
\bm{p} = \left[ \sum_j{(\bm{I} - \bm{R}_{oj}\bm{R}_{oj}^{T})}\right]^{-1}\left[ \sum_j{(\bm{I} - \bm{R}_{oj}\bm{R}_{oj}^{T})\bm{x}_j}\right].
\label{Eq:RayTracing2}
\end{equation}
The detailed derivation refers to  \cite{Szeliski10}.
The depth of each point $p$ is its $z$ component. \Fref{Fig:2Stereo} is an example of the depth map in water drop image.

%\paragraph{Structure from Motion}
%\shaodi{Optional, will update this part once I finish the experiments.}

\subsection{Rectification of Water Drop Image}
%\paragraph{Geometric Rectification}

%\begin{figure*}[tb]
%	\includegraphics[width=\linewidth]{Fig/Dewarp.pdf}
%	\caption{Rectified water drop images.}
%	\label{Fig:Dewarp}
%\end{figure*}

\begin{figure*}[tb]
	\includegraphics[width=\linewidth]{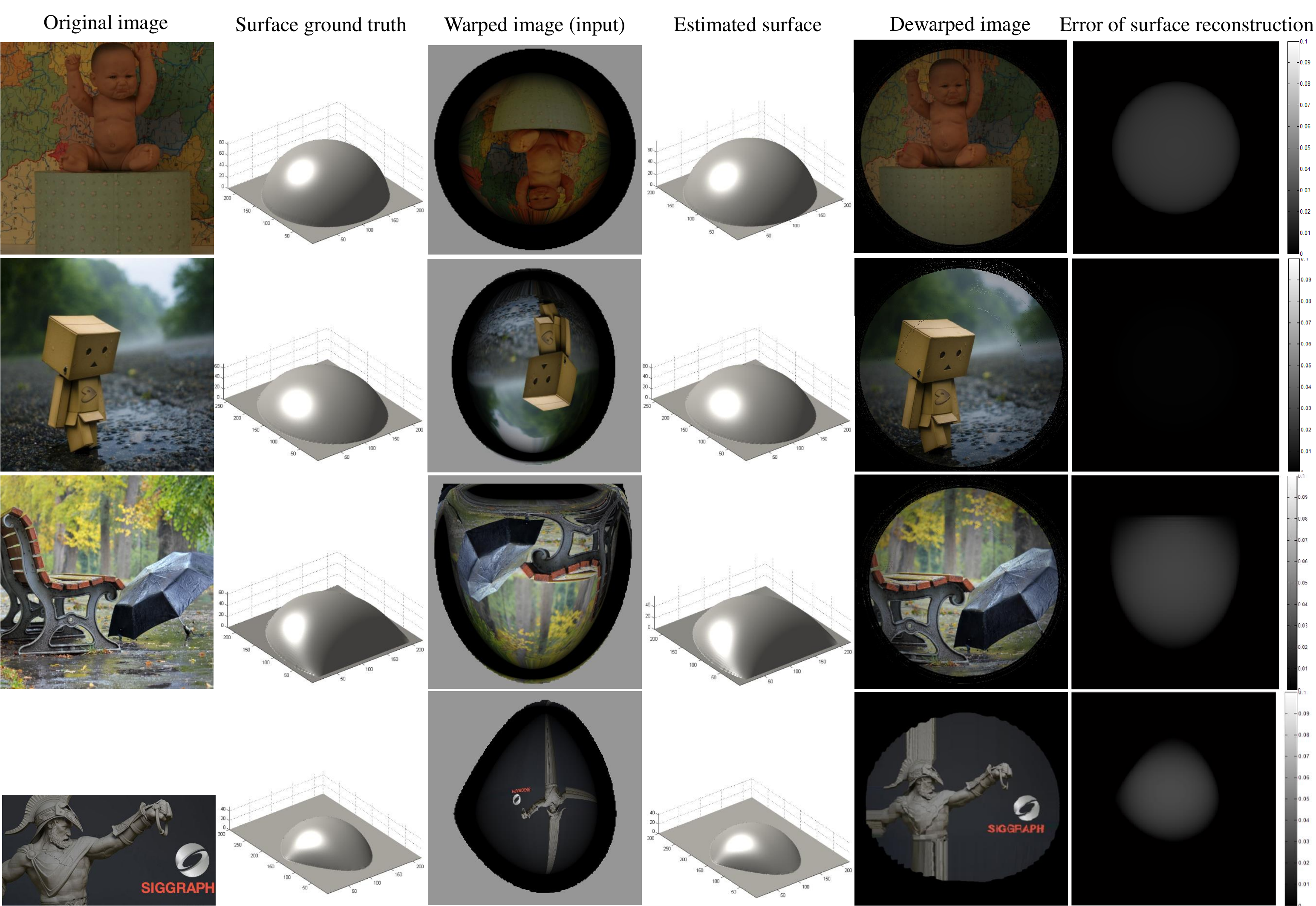}
	\caption{Quantitative evaluation of water surface reconstruction and rectification.}
	\label{Fig:3DSyn}
\end{figure*}

\begin{figure*}[tb]
	\includegraphics[width=\linewidth]{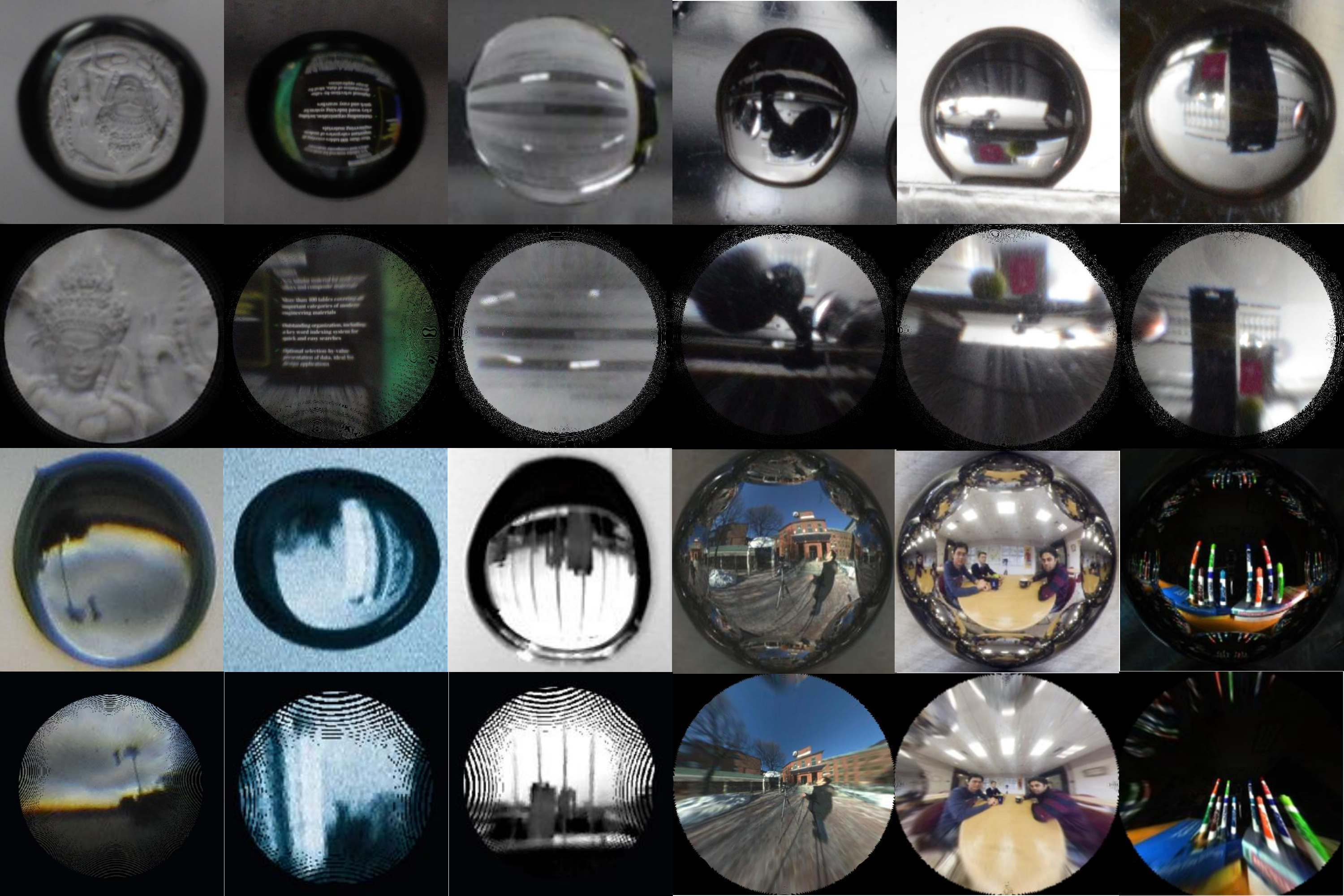}
	\caption{Rectification of real water images.} The first two row are water-drop images taken by ourselves. The first three columns in Row 3 and 4 are downloaded from the Internet. And the last three images are taken by sperical mirrors by \cite{Taguchi10}.
	There is slant between the background and the water drop in some data, however the rectified image is not necessary to be rectangles.
	\label{Fig:3DReal}
\end{figure*}

Having estimated the depth map on each water drop, we  unwarp the distorted water drop image. Referring to the pinhole camera model (\fref{Fig:PinHole}), for each water drop image, a space point $P_r$ with projection at $P_{ri}$ is projected to $P_e$. \Fref{Fig:3DSyn} and \Fref{Fig:3DReal} shows results of the rectified water drop images.

%\paragraph{Photometric Rectification}
According to \eref{Eq:Ts} and \eref{Eq:Tp}, with the water drop geometry obtained, we can compensate the brightness values according to the refractive coefficient $\mathcal{T}$. \Fref{Fig:Illuminance} shows an example of the brightness compensation. 

\begin{figure*}[tbh]
	\includegraphics[width=\linewidth]{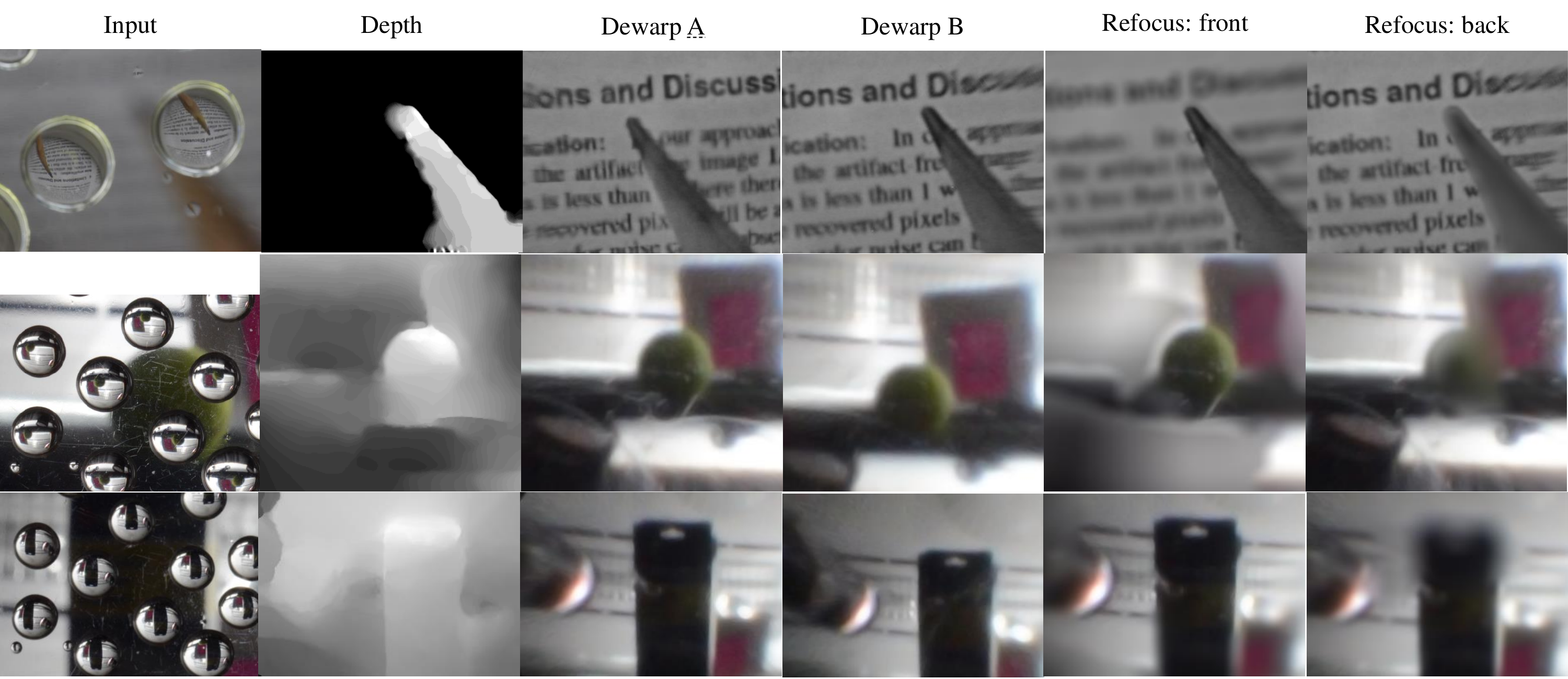}
	\caption{Stereo using real waterdrop images.} The stereo uses 2-4 waterdrops, however only two of the dewarp results are shown.
	\label{Fig:StereoR}
\end{figure*}

%%%%%%%%%%%%%%%%%%%%%%%%%%%%%%%%%%%%%%%
%Section
%%%%%%%%%%%%%%%%%%%%%%%%%%%%%%%%%%%%%%%
\section{Experiments and Analysis}

We conduct experiments using both synthetic data and real data to examine and analyze the performance of our method. In the experiment, we evaluate the estimated 3D shape of the water drops, and the depth estimation. Without loss of generality, our method can also be used to handle axial mirror/lens models \cite{Taguchi10}.

\subsection{3D Shape Reconstruction and Image Rectification}

To evaluate the accuracy of the 3D shape of water drops, we utilize synthetic data. We cannot use real data, since automatic 3D acquisition systems, such as a laser range finder, cannot be used to estimate the 3D of water.
We use real images for the evaluation of the image rectification. Some of the real images are taken by ourselves and some are downloaded from the Internet.

%\paragraph{Quantitative Evaluation Using Synthetic Data}
\Fref{Fig:3DSyn} shows the generated synthetic water drops with a variety of boundaries. A quantitative evaluation is performed by comparing our estimation with the ground truth 3D shape. The error is normalized to the percentage of the scale of the water drops. As one can observe, the reconstruction error is less than 3\% even for the most irregular water drops.

%\paragraph{Qualitative Evaluation Using Real Water Drops}
\Fref{Fig:3DReal} shows a collection of the rectified water drop images from real data. The input image is cropped for better visualization, yet the camera center is not at the cropped image center. 

%\paragraph{Qualitative Evaluation Using Real Spherical Mirror}
Without loss of generality, our proposed method can also be used for axial models, which is considered as a specific case when the water drop is exactly radially symmetric. The last two rows of \fref{Fig:3DReal} show the results on spherical mirrors \cite{Taguchi10}. Because dark band estimation is not applicable on mirrors, we specify the volume parameter $\alpha = 0.40$ for the mirrors. And refraction is changed to reflection in raytracing.

%\paragraph{Performance}
We implemented our method in Matlab and measured the computational time without parallelization. For water 3D shape estimation, the time varies depending on the water drop volume and the mesh resolution. \Tref{T:Time1} shows the computation time of varying volume and fixed mesh resolution. And \Tref{T:Time2} shows the time of varying mesh resolution. At typical case, the resolution of mesh is set to 200$\times$200 and the reconstruction time is about 10s. 
Note that, because each of the water drop reconstruction are performed separately, we can simply parallelize each of the tasks. Thus, the overall computation time does not increase with the number of water drops.

\begin{table}[tb]
	\caption{Computation time for water drop 3D reconstruction with varying volume.}  
	\includegraphics[width=\linewidth]{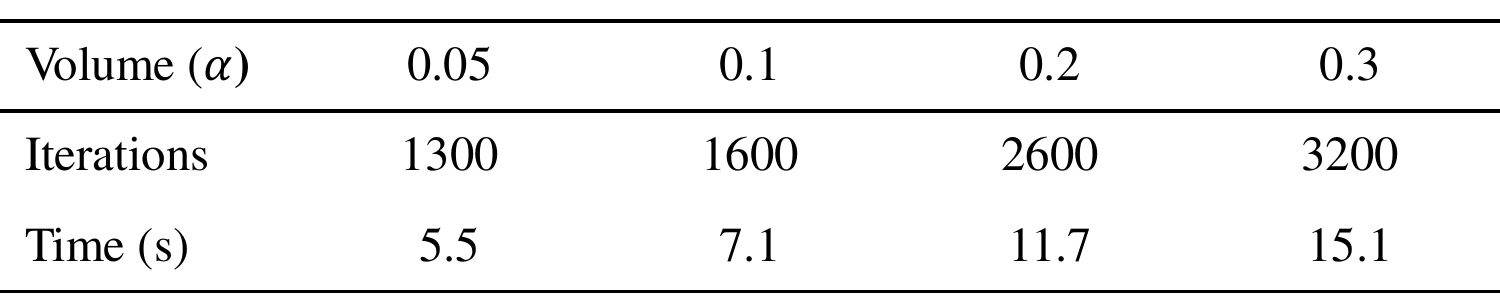}
	The mesh resolution is fixed to 200$\times$200.
	\label{T:Time1}
\end{table}

\begin{table}[tb]
	\caption{Computation time for water drop 3D reconstruction with varying mesh resolution.}  
	\includegraphics[width=\linewidth]{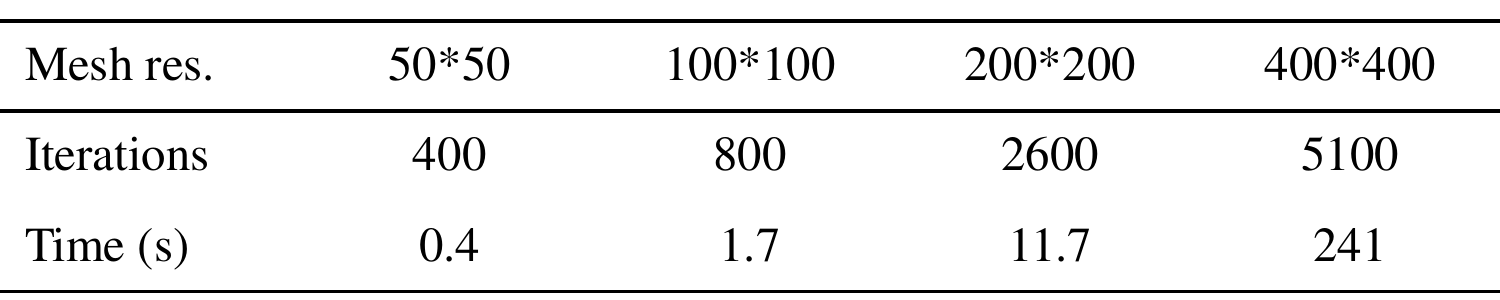}
	The volume is fixed to $\alpha = 0.2$.
	\label{T:Time2}
\end{table}

\subsection{Depth Estimation}

We use both synthetic and real water drop data to demonstrate our stereo method. Furthermore, Our non-parametric, non-axial method could be applied to axial-mirror/lens model as well.

\Fref{Fig:2Stereo} shows the generated synthetic data from the Middlebury data set. As can be seen, the depth estimation result highly resembles the ground truth with only errors occur at object's boundaries.

%\paragraph{Qualitative Evaluation Using Real Water Drops}
The result on the real waterdrop images are shown in the first 4 rows of \fref{Fig:StereoR}. For the first row, the data is taken using a micro-lens, and the resolution for each waterdrop is more than 600 pixels. The second and third data is taken by normal commercial lens, the resolution for each waterdrop is about 200-300 pixels. 
As shown, our proposed method can generally recover the depth structure. However, we find the bottleneck of our method is in finding the corresponding points between water drop images. Since, when the camera is zoomed-in to focus on the details of water drops, the sensor noise and the dust on the plate is no longer negligible, which adversely affect the accuracy and stability of both sparse and dense corresponding methods.

\begin{figure}[t]
	\includegraphics[width=\linewidth]{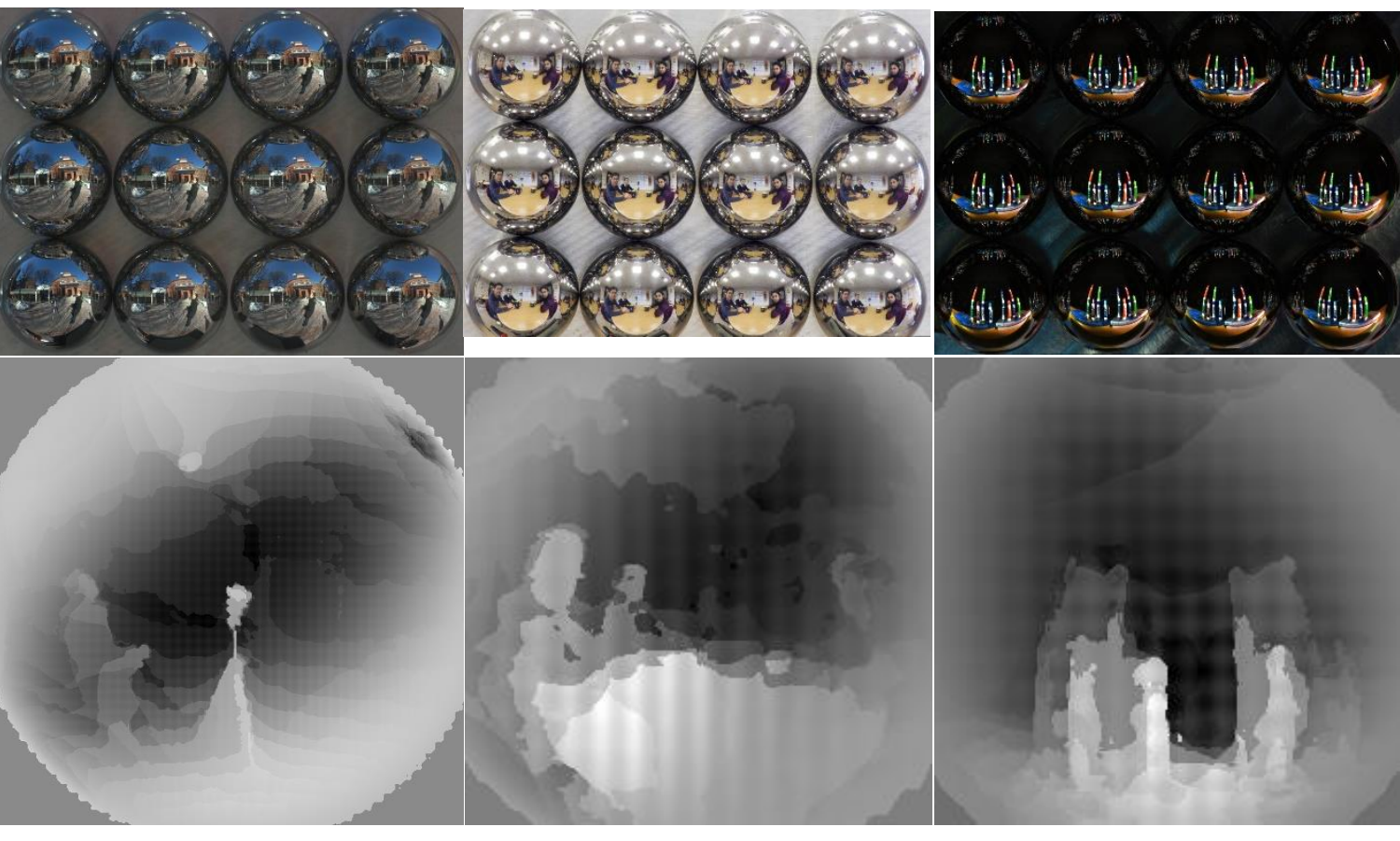}
	\caption{Stereo using axial mirror images.} Our non-parametric, non-axial method can be applied to axial data as well. The first column is the result using the original data provided by \cite{Taguchi10}, the second and third column are based on low-resolution data, because the original data is not provided.
	\label{Fig:AxialCone}
\end{figure}

As mentioned previously, our non-parametric, non-axial  method can be applied to axial-mirror/lens model.  \fref{Fig:AxialCone} shows our stereo estimation results. Since the image has sufficiently high resolution and less noise, the dense matching is significantly stable, and consequently it enables us to estimate the depth more accurately.

%%%%%%%%%%%%%%%%%%%%%%%%%%%%%%%%%%%%%%%
%Section
%%%%%%%%%%%%%%%%%%%%%%%%%%%%%%%%%%%%%%%
\section{Discussion and Conclusion}

In this paper, we had exploited the depth reconstruction from water drops. In our pipeline, there are three key steps: the water-drop 3D shape reconstruction, depth estimation using stereo, and water-drop image rectification. All of these are done using a single image. 
We evaluated our method, and it shows that the method works effectively for both synthetic and real images. Nevertheless, there are still some limitations in it.
One of the limitation is the common perspective camera and lens cannot obtain  high resolution image of water drops. Which degraded the overall performance of the depth estimation, specifically, the sparse/dense correspondence quality is degraded because of the low-resolution images.
For future works, we are considering improving the image quality. Furthermore,  we will consider simultaneous estimation of the waterdrop geometry and  depth.

\bibliographystyle{acmsiggraph}
%\nocite{*}
\bibliography{YSDbib}

\begin{thebibliography}{\protect\citename{Swaminathan et~al\mbox{.} }2001}

\bibitem[\protect\citename{Agrawal and Ramalingam }2013]{Agrawal13}
{\sc Agrawal, A., and Ramalingam, S.}
\newblock 2013.
\newblock Single image calibration of multi-axial imaging systems.
\newblock In {\em Computer Vision and Pattern Recognition (CVPR), 2013 IEEE
  Conference on}, IEEE, 1399--1406.

\bibitem[\protect\citename{Baker and Nayar }1999]{Baker99}
{\sc Baker, S., and Nayar, S.~K.}
\newblock 1999.
\newblock A theory of single-viewpoint catadioptric image formation.
\newblock {\em International Journal of Computer Vision 35}, 2, 175--196.

\bibitem[\protect\citename{Favaro and Soatto }2005]{Favaro05}
{\sc Favaro, P., and Soatto, S.}
\newblock 2005.
\newblock A geometric approach to shape from defocus.
\newblock {\em Pattern Analysis and Machine Intelligence, IEEE Transactions on
  27}, 3, 406--417.

\bibitem[\protect\citename{Feynman et~al\mbox{.} }2013]{Feynman13}
{\sc Feynman, R.~P., Leighton, R.~B., and Sands, M.}
\newblock 2013.
\newblock {\em The Feynman Lectures on Physics, Desktop Edition Volume I},
  vol.~1.
\newblock Basic Books.

\bibitem[\protect\citename{Garg and Nayar }2007]{Garg07}
{\sc Garg, K., and Nayar, S.}
\newblock 2007.
\newblock Vision and rain.
\newblock {\em International Journal of Computer Vision 75}, 1, 3--27.

\bibitem[\protect\citename{Hartley and Zisserman }2004]{Hartley04}
{\sc Hartley, R.~I., and Zisserman, A.}
\newblock 2004.
\newblock {\em Multiple View Geometry in Computer Vision}, second~ed.
\newblock Cambridge University Press, ISBN: 0521540518.

\bibitem[\protect\citename{Hassner and Basri }2006]{Hassner06}
{\sc Hassner, T., and Basri, R.}
\newblock 2006.
\newblock Example based 3d reconstruction from single 2d images.
\newblock In {\em Computer Vision and Pattern Recognition Workshop, 2006.
  CVPRW'06. Conference on}, IEEE, 15--15.

\bibitem[\protect\citename{Horry et~al\mbox{.} }1997]{Horry97}
{\sc Horry, Y., Anjyo, K.-I., and Arai, K.}
\newblock 1997.
\newblock Tour into the picture: using a spidery mesh interface to make
  animation from a single image.
\newblock In {\em Proceedings of the 24th annual conference on Computer
  graphics and interactive techniques}, ACM Press/Addison-Wesley Publishing
  Co., 225--232.

\bibitem[\protect\citename{Ikeuchi and Horn }1981]{Ikeuchi81}
{\sc Ikeuchi, K., and Horn, B.~K.}
\newblock 1981.
\newblock Numerical shape from shading and occluding boundaries.
\newblock {\em Artificial intelligence 17}, 1, 141--184.

\bibitem[\protect\citename{Joshi and Carr }2008]{Joshi08}
{\sc Joshi, P., and Carr, N.~A.}
\newblock 2008.
\newblock Repouss{\'e}: automatic inflation of 2d artwork.
\newblock In {\em Proceedings of the Fifth Eurographics conference on
  Sketch-Based Interfaces and Modeling}, Eurographics Association, 49--55.

\bibitem[\protect\citename{Kanaev et~al\mbox{.} }2012]{Kanaev12}
{\sc Kanaev, A.~V., Hou, W., Woods, S., and Smith, L.~N.}
\newblock 2012.
\newblock Restoration of turbulence degraded underwater images.
\newblock {\em Optical Engineering 51}, 5, 057007--1.

\bibitem[\protect\citename{Levoy et~al\mbox{.} }2004]{Levoy04}
{\sc Levoy, M., Chen, B., Vaish, V., Horowitz, M., McDowall, I., and Bolas, M.}
\newblock 2004.
\newblock Synthetic aperture confocal imaging.
\newblock In {\em ACM Transactions on Graphics (TOG)}, vol.~23, ACM, 825--834.

\bibitem[\protect\citename{Malik and Rosenholtz }1997]{Malik97}
{\sc Malik, J., and Rosenholtz, R.}
\newblock 1997.
\newblock Computing local surface orientation and shape from texture for curved
  surfaces.
\newblock {\em International journal of computer vision 23}, 2, 149--168.

\bibitem[\protect\citename{Morris }2004]{Morris04}
{\sc Morris, N. J.~W.}
\newblock 2004.
\newblock {\em Image-based water surface reconstruction with refractive
  stereo}.
\newblock PhD thesis, University of Toronto.

\bibitem[\protect\citename{Oreifej et~al\mbox{.} }2011]{Oreifej11}
{\sc Oreifej, O., Shu, G., Pace, T., and Shah, M.}
\newblock 2011.
\newblock A two-stage reconstruction approach for seeing through water.
\newblock In {\em Computer Vision and Pattern Recognition (CVPR), 2011 IEEE
  Conference on}, IEEE, 1153--1160.

\bibitem[\protect\citename{Oswald et~al\mbox{.} }2012]{Oswald12}
{\sc Oswald, M.~R., Toppe, E., and Cremers, D.}
\newblock 2012.
\newblock Fast and globally optimal single view reconstruction of curved
  objects.
\newblock In {\em Computer Vision and Pattern Recognition (CVPR), 2012 IEEE
  Conference on}, IEEE, 534--541.

\bibitem[\protect\citename{Prasad and Fitzgibbon }2006]{Prasad06}
{\sc Prasad, M., and Fitzgibbon, A.}
\newblock 2006.
\newblock Single view reconstruction of curved surfaces.
\newblock In {\em Computer Vision and Pattern Recognition, 2006 IEEE Computer
  Society Conference on}, vol.~2, IEEE, 1345--1354.

\bibitem[\protect\citename{Ramalingam et~al\mbox{.} }2006]{Ramalingam06}
{\sc Ramalingam, S., Sturm, P., and Lodha, S.~K.}
\newblock 2006.
\newblock Theory and calibration for axial cameras.
\newblock In {\em Computer Vision--ACCV 2006}. Springer, 704--713.

\bibitem[\protect\citename{Roser and Geiger }2009]{Roser09}
{\sc Roser, M., and Geiger, A.}
\newblock 2009.
\newblock Video-based raindrop detection for improved image registration.
\newblock {\em IEEE 12th International Conference on Computer Vision Workshops
  (ICCV Workshops)\/}.

\bibitem[\protect\citename{Roser et~al\mbox{.} }2010]{Roser10}
{\sc Roser, M., Kurz, J., and Geiger, A.}
\newblock 2010.
\newblock Realistic modeling of water droplets for monocular adherent raindrop
  recognition using bezier curves.
\newblock {\em Asian Conference on Computer Vision\/}.

\bibitem[\protect\citename{Swaminathan et~al\mbox{.} }2001]{Swaminathan01}
{\sc Swaminathan, R., Grossberg, M.~D., and Nayar, S.~K.}
\newblock 2001.
\newblock Caustics of catadioptric cameras.
\newblock In {\em Computer Vision, 2001. ICCV 2001. Proceedings. Eighth IEEE
  International Conference on}, vol.~2, IEEE, 2--9.

\bibitem[\protect\citename{Szeliski }2010]{Szeliski10}
{\sc Szeliski, R.}
\newblock 2010.
\newblock {\em Computer vision: algorithms and applications}.
\newblock Springer Science \& Business Media.

\bibitem[\protect\citename{Taguchi et~al\mbox{.} }2010]{Taguchi10}
{\sc Taguchi, Y., Agrawal, A., Veeraraghavan, A., Ramalingam, S., and Raskar,
  R.}
\newblock 2010.
\newblock Axial-cones: modeling spherical catadioptric cameras for wide-angle
  light field rendering.
\newblock {\em ACM Transactions on Graphics-TOG 29}, 6, 172.

\bibitem[\protect\citename{Terzopoulos et~al\mbox{.} }1988]{Terzopoulos88}
{\sc Terzopoulos, D., Witkin, A., and Kass, M.}
\newblock 1988.
\newblock Symmetry-seeking models and 3d object reconstruction.
\newblock {\em International Journal of Computer Vision 1}, 3, 211--221.

\bibitem[\protect\citename{Tian and Narasimhan }2009]{Tian09}
{\sc Tian, Y., and Narasimhan, S.~G.}
\newblock 2009.
\newblock Seeing through water: Image restoration using model-based tracking.
\newblock In {\em Computer Vision, 2009 IEEE 12th International Conference on},
  IEEE, 2303--2310.

\bibitem[\protect\citename{Vicente and Agapito }2013]{Vicente13}
{\sc Vicente, S., and Agapito, L.}
\newblock 2013.
\newblock Balloon shapes: reconstructing and deforming objects with volume from
  images.
\newblock In {\em 3D Vision-3DV 2013, 2013 International Conference on}, IEEE,
  223--230.

\bibitem[\protect\citename{Xu et~al\mbox{.} }2012]{Xu12}
{\sc Xu, L., Jia, J., and Matsushita, Y.}
\newblock 2012.
\newblock Motion detail preserving optical flow estimation.
\newblock {\em Pattern Analysis and Machine Intelligence, IEEE Transactions on
  34}, 9, 1744--1757.

\bibitem[\protect\citename{You et~al\mbox{.} }2013]{You13}
{\sc You, S., Tan, R.~T., Kawakami, R., and Ikeuchi, K.}
\newblock 2013.
\newblock Adherent raindrop detection and removal in video.
\newblock {\em IEEE Computer Society Conference on Computer Vision and Pattern
  Recognition (CVPR)\/}.

\bibitem[\protect\citename{You et~al\mbox{.} }2015]{You15}
{\sc You, S., Tan, R., Kawakami, R., Mukaigawa, Y., and Ikeuchi, K.}
\newblock 2015.
\newblock Adherent raindrop modeling, detection and removal in video.
\newblock {\em Pattern Analysis and Machine Intelligence, IEEE Transactions
  on\/}.

\bibitem[\protect\citename{Zorich and Cooke }2004]{Zorich04}
{\sc Zorich, V., and Cooke, R.}
\newblock 2004.
\newblock Mathematical analysis.
\newblock {\em Springer\/}.

\end{thebibliography}

%%%%%%%%%%%%%%%%%%%%%%%%%%%%%%%%%%%%%%%
%Section
%%%%%%%%%%%%%%%%%%%%%%%%%%%%%%%%%%%%%%%
\section*{Appendix A: Equivalent Camera Position for Flat Air-Water Refraction}

Here we prove that we can further remove the refraction from water to air in \fref{Fig:Brewster}.a by moving the camera to its approximated equivalent place at $C'$, as illustrated in \fref{Fig:Brewster}.b.

We assume the camera position is $\bm{C} = (0, 0, z_c)^\top$ and the flat plate is $z = 0$. Given a point on the plate $\bm{x} = (x, y, 0)^\top$ where the refraction happens, the orientation of the incidence angle is:
\begin{equation}
\bm{R}_i = \frac{\bm{x} - \bm{C}}{\parallel\bm{x} - \bm{C}\parallel} = \frac{(x, y, -z_c)^\top}{\sqrt{x^2 + y^2 + z_c^2}}
\label{Eq:A1}
\end{equation}
The parallel and orthogonal components to the surface normal $N = (0, 0, 1)^\top$ are:
\begin{eqnarray}
\begin{split}
\bm{R}_{i\parallel} =& (\bm{R}_i^\top\bm{N})\bm{N} = \frac{(0, 0, -z_c)^\top}{\sqrt{x^2 + y^2 + z_c^2}}
\\
\bm{R}_{i\top} =& \bm{R}_i - \bm{R}_{i\parallel} = \frac{(x, y, 0)^\top}{\sqrt{x^2 + y^2 + z_c^2}}.
\end{split}
\label{Eq:A2}
\end{eqnarray}

The orientation of angle of refraction, denoted as $\bm{R}_o$, can be obtained according to Snell's law:
\begin{eqnarray}
\bm{R}_{o\top} &=& \frac{n_a}{n_w} \bm{R}_{i\top} = \frac{(x, y, 0)^\top}{\frac{n_w}{n_a}\sqrt{x^2 + y^2 + z_c^2}}
\label{Eq:A3}
\end{eqnarray}
Considering $\bm{R}_o$ is normalized: $\parallel\bm{R}_{o}\parallel=\parallel\bm{R}_{o\top}+\bm{R}_{o\parallel}\parallel=1$, and thus:
\begin{equation}
\bm{R}_{o}= \frac{\bm{R}'_o}{\parallel\bm{R}'_o\parallel}=\frac{(x, y, \frac{n_w}{n_a}z_c\sqrt{1 + \frac{n_w^2 - n_a^2}{n_w^2}\frac{x^2+y^2}{z_c^2}})^\top}{\parallel\bm{R}'_{o}\parallel}.
\label{Eq:A4}
\end{equation}
Hence, the equivalent camera position is: \hbox{$\bm{C}' = (0, 0, \frac{n_w}{n_a}z_c\sqrt{1 + \frac{n_w^2 - n_a^2}{n_w^2}\frac{x^2+y^2}{z_c^2}})^\top$}
when the incidence angle is close to the optic axis of the camera, \ie, $x, y \ll z_c$, we can have the approximation that the equivalent camera position \hbox{$\bm{C}' = (0, 0, \frac{n_w}{n_a}z_c)^\top$}.

%%%%%%%%%%%%%%%%%%%%%%%%%%%%%%%%%%%%%%%
%Section
%%%%%%%%%%%%%%%%%%%%%%%%%%%%%%%%%%%%%%%
\section*{Appendix B: Detailed Implementation of Water-drop Multiple View Stereo}

Once the geometry of each water drop is obtained, we perform multiple view stereo to estimate depth. We propose a raytracing based triangulation. As illustrated in \fref{Fig:StereoModel}, for two corresponding points $\bm{x}_1$ and $\bm{x}_2$ and their rays of refractions, the triangulation aims to find the position of point $\bm{p}$ which minimizes the Euclidean distance to all the rays. This idea can be directly extended to more than two water drops.

There are 3 main steps in our multiple view steres: (1) Inverse raytracing, (2)  Corresponding points for different water drops, and (3) Triangulation.

\paragraph{Inverse Raytracing}
The goal of the inverse raytracing is to find the orientation of the ray of refraction. We call it inverse raytracing because we assume the ray is originated from the camera, refracted by the water drops and arrives at the objects. 

As illustrated in \fref{Fig:StereoModel}, we show the inverse raytracing on the left water drop. Without loss of generality, we assume the camera position is $\bm{C} = (0, 0, z_c)^\top$ and the flat plate is $z = 0$, and the image-plane has corresponding pixels with the flat plane using rotation and scaling. According to Appendix A, the equivalent camera position is \hbox{$\bm{C}' = (0, 0, z'_c)^\top = (0, 0, \frac{n_w}{n_a}z_c\sqrt{1 + \frac{n_w^2 - n_a^2}{n_w^2}\frac{x^2+y^2}{z_c^2}})^\top$}.

For a pixel on the image plane, with a corresponding point on the flat plate $\bar{\bm{x}} = (\bar{x}, \bar{y}, \bar{z})^\top$, we can find the refraction location, denoted as $\bm{x} = (x, y, z) ^\top$, by using the constraints:
\begin{eqnarray}
\frac{x}{\bar{x}} = \frac{y}{\bar{y}} = \frac{z - z'_c}{\bar{z} - z'_c}.
\end{eqnarray}
In practice, because finding the intersection between a flat plane and a line is easier than finding the intersection between a line and a curved surface, we specify $\bm{x}$ and find the corresponding pixel $\bar{\bm{x}}$.

At point $\bm{x}$, the angle of incidence is:
\begin{equation}
\bm{R}_i = \frac{\bm{x} - \bm{C}'}{\parallel\bm{x} - \bm{C}'\parallel} = \frac{(x, y, z - z'_c)^T}{\parallel\bm{x} - \bm{C}'\parallel}.
\end{equation}
The surface normal is obtained according to the water drop geometry:
\begin{eqnarray}
\begin{split}
\bm{N} =& \frac{\bm{N}'}{\parallel\bm{N}'\parallel}
\\
\bm{N}' =& (\frac{\partial{z}}{\partial{x}}, \frac{\partial{z}}{\partial{z}}, 1)^\top.
\end{split}
\end{eqnarray}.

Then, the orientation of ray of refraction $\bm{R}_o$ is obtained according to Snell's law:
\begin{eqnarray}
\begin{split}
\bm{R}_{i\parallel} =& (\bm{R}_i^\top\bm{N})\bm{N}
\\
\bm{R}_{i\top} =& \bm{R}_i - \bm{R}_{i\parallel}
\\
\bm{R}_{o\top} =& \frac{n_w}{n_a} \bm{R}_{i\top}
\\
\bm{R}_{o} =& \bm{R}_{o\top} + \bm{R}_{o\parallel}
\\
\parallel\bm{R}_o\parallel = & 1.
\end{split}
\end{eqnarray}

\paragraph{Finding Correspondence Between Water Drops}
Finding the corresponding pixels between different warped water drop images is a challenging task. Compared to the normal cameras, the distortion between water drop images is significantly worse. Morever, unlike spherical mirrors/lenses where all the mirror/lenses share the same distortion, each water drop has its own distortion. 

To solve this problem, we try to find the corresponding pixels on the angular-dewarped images. Note that, since the depth of image is not yet obtained, we cannot accurately dewarped the image. Thus, we dewarp the images solely according to the angle of refraction. Nevertheless, we find the angular-dewarping can signficantly recover the images.

As introduced in the inverse ray-tracing, for a pixel $\bar{\bm{x}}$, with ray of refraction \hbox{$\bm{R}_o = (r_{ox}, r_{oy}, r_{oz})^\top$}, we project the pixel to 
\begin{eqnarray}
(\frac{r_{ox}}{r_{oz}}, \frac{r_{oy}}{ r_{oz}})^\top .
\end{eqnarray}

Because the water drop surface is smooth and convex, implying the Jacobian on the surface is always positive, it  means the angular mapping is one-to-one \cite{Zorich04}. Thus, we can map back the dewarped corresponding pixels to the warped image.

The third and fourth columns of \Fref{Fig:2Stereo} show examples of the angular dewarping results and the dense correspondence on the dewarped images. Specifically, we use \cite{Xu12} for the dense correspondence estimation.

\paragraph{Triangulation}
Now we can perform the classical triangulation as illustrated in \fref{Fig:StereoModel}. Given a set of corresponding pixels on each water drop $\bm{x}_j$, we can obtain its outbound ray of refraction:
\begin{equation}
\bm{x}_j + \alpha\bm{R}_{oj}, \alpha\in\mathbb{R}, j = 1, 2, \cdots .
\end{equation}
The triangulation's goal is to find the position of point $\bm{p}$, which minimizes the Euclidean distance to all the rays \cite{Hartley04}:
\begin{equation} 
\bm{p} = \left[ \sum_j{(\bm{I} - \bm{R}_{oj}\bm{R}_{oj}^{T})}\right]^{-1}\left[ \sum_j{(\bm{I} - \bm{R}_{oj}\bm{R}_{oj}^{T})\bm{x}_j}\right].
\end{equation}
The depth is the $z$ component of $\bm{p}$.

\end{document}